\documentclass[final]{agujournal2019}
\usepackage{url} 
\usepackage{lineno}
\usepackage{soul}
\usepackage{amsmath}
\usepackage{amssymb}
\usepackage{multirow}
\usepackage{xurl}
\usepackage{booktabs}
\usepackage{float}
\draftfalse

\journalname{JGR: Machine Learning and Computation}

\makeatletter
\def\@oddhead{}
\def\@evenhead{}
\makeatother

\begin{document}

%
%


\title{PACT: Peak-Aware Cross-Attention Graph Transformers for Efficient Storm-Surge Emulation}

%
%




\authors{Zesheng Liu\affil{1}, Doyup Kwon\affil{2}, Ning Lin\affil{2}, Maryam Rahnemoonfar\affil{1,3}}


\affiliation{1}{Department of Computer Science and Engineering, Lehigh University}
\affiliation{2}{Department of Civil and Environmental Engineering, Princeton University}
\affiliation{3}{Department of Civil and Environmental Engineering, Lehigh University}




\correspondingauthor{Maryam Rahnemoonfar}{maryam@lehigh.edu}



\begin{keypoints}
\item GNN-based emulation is posed as a computationally efficient tool for long-term storm surge time series projection.
\item A station conditioned graph transformer learns where and when gridded forcing matters for each location.
\item Peak aware learning improves the prediction of rare extremes and supports transfer between different forcing sources.
\end{keypoints}


%
%

%
%


\begin{abstract}
Accurate and efficient storm-surge emulation is essential for coastal hazard assessment, yet high-fidelity hydrodynamic models remain too expensive for large scenario ensembles and rapid evaluation under heterogeneous climate forcings. We present PACT, a peak-aware cross-attention graph transformer for efficient station-level storm-surge prediction from atmospheric forcing fields. PACT represents each forcing patch as a graph, encodes spatial structure with GraphSAGE, and uses a learned station query to aggregate node information through cross-attention rather than uniform pooling. A Transformer encoder models temporal dependence across the forcing history, and a horizon-query decoder generates lead-specific forecasts from a shared temporal memory. To better capture extreme events, we introduce a peak-aware learning strategy that couples a lightweight auxiliary peak-aware head with a tailored training objective, including a tail-focused loss on peak-dominated samples and a horizon-wise slope regularizer to encourage coherent multi-step evolution. Across multiple tide-gauge stations along the US Northeast coast, PACT outperforms a strong spatio-temporal graph neural network baseline in both RMSE and MAE. Diagnostics show improved peak fidelity and tail preservation for reanalysis and most CMIP6 datasets. PACT is also computationally efficient, requiring about 3.5~s to generate a full winter-season surge trajectory for one year after training. Under distribution shift across five CMIP6 forcings, PACT transfers well within the CMIP6 family but degrades markedly when transferring from reanalysis to climate-model forcings, highlighting a persistent reanalysis--GCM gap.
\end{abstract}

\section*{Plain Language Summary}
Storm surge is one of the main causes of coastal flooding during severe storms, but accurate surge prediction usually depends on large numerical models that can take several hours to run. This makes it difficult to produce fast forecast or explore different future stor scenarios. In this study, we developed a much faster machine learning model that predicts water level timeseries at coastal stations from spatial-temporal weather conditions such as wind velocities and air pressure. Our model is designed to focus on the parts of the weather field that matter most for each station, learn how conditions evolve over time, and better capture rare but dangerous extreme storm surge events. We tested our model on both historical and climate model datasets. The results show that it can make predictions with high accuracy, improve performance for extreme surge events compared with simpler approaches, and can generate a full winter season surge level predictions in only a few seconds instead of hours.


%
%

%


%
%
%
%



\section{Introduction}
\label{sec:intro}

Storm surge is among the most damaging coastal hazards, contributing substantially to coastal flooding during both tropical cyclones (TCs) and extratropical cyclones (ETCs)~\cite{Storm_Surge_Hazards, Bertin_StormSurge}. Storm surge is defined as the abnormal rise of water level above the astronomical tide, driven primarily by wind stress and atmospheric pressure and modulated by coastal geometry, bathymetry, and background ocean conditions~\cite{Qian_2024,Islam2021}. Because surge response can vary sharply across nearby locations, station-level tide-gauge time series are a key simulation target for forecasting, emergency management, risk assessment, and long-term coastal planning.

High-fidelity hydrodynamic models such as ADCIRC and SWAN+ADCIRC remain the standard for resolving storm-surge dynamics over complex coastal domains ~\cite{adcirc, DIETRICH_2011_SWAN_ADCIRC,Dietrich_2012_SWAN_ADCIRC}. These models solve the governing equations of fluid flow to generate spatial-temporal records of water elevation, and they are widely used for operational forecasting, scenario analysis, and coastal hazard studies~\cite{sebastian2014,glahn2009,fleming2008,begmohammadi2025,zhu2026,dinapoli2025}. However, their computational cost is substantial, particularly for the large ensembles of simulations required for probabilistic forecasting, uncertainty quantification, and climate-impact studies~\cite{lin2019,LEE2021104024,Pachev_2023}. This computational bottleneck has motivated growing interest in data-driven surrogate models that learn from observations or high-fidelity numerical simulations and can then approximate surge behavior at much lower cost~\cite{butler2012,pachev2023}. 

Most existing machine-learning surge surrogates have been developed for TC-focused settings and often rely on time series of storm parameters such as track, size, translation speed, and intensity. For example, \citeA{valle2021} and \citeA{lockwood2022} used artificial neural networks (ANNs) to predict storm-surge time series from these TC  parameters, with training data generated from synthetic storms driving ADCIRC. Similarly, \citeA{KYPRIOTI2023104231} developed Gaussian-process surrogate models for storm-surge time-series emulation and \citeA{Adeli2023} proposed an encoder-decoder convolutional long short-term memory (ConvLSTM)  surrogate for multi-node surge time series. These approaches demonstrate that learning-based models can efficiently approximate numerical simulations when representing meteorological forcing by a small set of parameters.

Other studies simplify the problem by predicting only the peak storm surge associated with each event~\cite{sahoo2019,hashemi2016,mohammad2024}. For example, \citeA{hashemi2016} trained an ANN to reproduce maximum storm surges generated by synthetic TCs in the northeastern United States, and \citeA{mohammad2024} evaluated annual probability distributions of water elevations estimated from ANN-predicted peak surges. Short-term station-level forecasting models have also been developed using convolutional neural network (CNN), LSTM, and ANN-based architectures, with lead times ranging from a few hours to about one day~\cite{XIE2023102179,Wang2023,KIM2019101871,w12092394,jmse10121980}. These approaches are valuable for their intended purposes, but peak-only models discard the temporal details of the surge hydrograph, while many short-lead models focus on event-centered or operational windows rather than continuous season-scale emulation across multiple years.

The TC focus of much prior work also leaves a gap for ETC-driven surge. ETCs are more frequent than TCs in many mid-latitude coastal regions~\cite{Sandeep_ETC_2024} and are responsible for a large fraction of major surge events in the northeastern United States; for example, as of 2017, ETCs caused 88 of the 100 largest storm events at The Battery, New York, and 91 of the 100 largest at Boston, Massachusetts~\cite{catalano2018}. Unlike TCs, which can often be represented by parametric descriptors, ETCs are larger systems whose impacts are tied to broad wind and pressure fields. Models designed for TC parameters are therefore not well suited to extracting the spatially distributed features that govern ETC surges. To address this difference in required data, some studies have moved beyond compact storm descriptors by using gridded meteorological fields from reanalysis or numerical models; \citeA{tadesse2020} and \citeA{tiggeloven2021} used atmospheric fields to train models to emulate surge time series at global coastal locations, while \citeA{zhu2026} focused specifically on ETC events in the Bohai Sea. There have also been efforts to apply meteorological fields to ML-based post-processing of numerical forecasts of storm surge ~\cite{TEDESCO2024102334}. However, many of these approaches do not focus explicitly on ETCs, include model augmentations for improved emulation of extreme surges, or investigate cross-dataset transfer capabilities. More fundamentally, standard deep learning architectures such as CNNs assume data defined on regular grids with local Euclidean neighborhoods, which makes them difficult to apply directly to the unstructured meshes commonly used in hydrodynamic and atmospheric models. Even when evenly gridded data are available, the dynamics governing storm surge often depend on non-local features, such as coastline changes or influences of forcing variables hundreds of km away. These dynamics may not be well captured by fixed convolutional kernels, motivating the need for an architecture that can more effectively represent irregular connectivities and relationships. 

Graph neural networks (GNNs) provide a natural modeling direction for storm-surge surrogates because they represent data as graphs, allowing for a more physically meaningful representation of spatial dependencies. This advantage has been demonstrated in related geophysical learning tasks, including internal ice-layer thickness prediction via graph neural networks~\cite{ Zalatan_IGARSS2023, Zalatan_ICIP2023,Zalatan_RadarConf23,liu2024multibranchspatiotemporalgraphneural, Liu_SPIE25}, physics-informed graph neural networks~\cite{Rahnemoonfar_PIML_IGARSS2024,liu2024learningspatiotemporalpatternspolar,Liu_PIML_RADAR25,liu2026kstemitknowledgeinformedspatiotemporalefficient}, graph transformers~\cite{Liu_GRIT_IGARSS2025, Liu_2025_GRIT_LP, Liu_STGRIT_ICIP25}, and GNN-based emulation on unstructured ice-sheet meshes~\cite{Koo_Rahnemoonfar_2025,Koo_Helheim, liu2025kangcncombiningkolmogorovarnoldnetwork}. Existing GNN-based storm-surge studies, however, remain relatively few and differ mainly in how graph structure is defined and what physical or statistical dependence the graph is intended to capture. Station-graph models such as  CSTGNN~\cite{JIANG2024104512} and StormNet~\cite{Nader_2026_StormNet} construct graphs over tide gauges to capture spatial dependence, using techniques such as information flow, correlation, and geographic proximity. These models demonstrate that encoding spatial relationships can improve forecasts across stations. Regional emulators such as Flo~\cite{kristensen2026flodatadrivenlimitedareastorm} learn the evolution of storm-surge fields from atmospheric forcing and hindcast data, showing that GNN-based regional emulators can reproduce large-scale surge dynamics with accuracy comparable to numerical results used for training. Finally, the GNN framework proposed by~\citeA{Movagha_2025} focuses on storm-surge inundation over extended geospatial domains. Rather than predicting continuous time series, it classifies the inundation state of many coastal nodes and introduces sparse connectivity, response-informed graph construction, and clustering-based decomposition to improve scalability across large domains. Together, these studies show that GNNs are promising for coastal hazard modeling, but also highlight that graph design is closely tied to the prediction object. This suggests that the evaluation of surrogate models requires first identifying the aspects of surge behavior that must be preserved.

Peak water levels are essential for threshold exceedance studies, design-oriented hazard assessment, and extreme-risk summaries, yet many applications require more than a single event maximum. Dynamic inundation modeling depends on the full hydrograph because flood extent and impacts are shaped by the timing, duration, and height of peaks~\cite{Dullaart_2023}. Multi-year continuous surge records are also needed for event cataloging, interannual risk characterization, and return-level estimation~\cite{Bernier_2006,return_level1,lin2019}. Large-scale surge reanalysis products similarly provide time series together with extreme-value summaries, reflecting the need to preserve temporal evolution rather than maxima alone~\cite{Muis2016}. 

Despite recent advances, existing GNN surrogates are not designed to meet these requirements. Station-graph models such as CSTGNN and StormNet focus on dependencies between observed gauges. While valuable for multi-station forecasting and forecast post-processing, these models do not directly learn station-conditioned representations from gridded atmospheric forcing over a shared region of interest. Regional field emulators such as Flo predict gridded residual water-level fields, but are not formulated for multi-horizon station-trajectory emulation. Inundation-state GNNs address wet/dry classification rather than continuous surge evolution. More broadly, architectural choices in existing models can limit their ability to capture key surge dynamics: compressing atmospheric forcing through fixed pooling can dilute the localized wind and pressure structures that drive surge at individual stations, while rigid temporal aggregation may miss event-dependent lag relationships between forcing and response. Finally, standard average-error losses tend to be dominated by the many low- and moderate-surge samples in the record, allowing a model to achieve good overall skill while underestimating rare peaks, misaligning peak timing, or producing unrealistic multi-step trajectories during rapidly evolving events. 

This setting is further complicated by differences in atmospheric forcing used for retrospective and future surge analysis. Reanalysis products such as the NCEP/NCAR Reanalysis~\cite{ncep} provide long, observationally constrained fields suitable for historical validation. Climate-model products meant for future projections, such as CMIP6~\cite{Eyring2016}, including ScenarioMIP experiments~\cite{ONeill_2016_ScenarioMIP}, differ in spatial resolution, temporal frequency, physical parameterizations, and storm climatologies. These differences are consequential: CMIP6-driven storm-surge projections can show systematic regional biases relative to each other~\cite{Muis_2023_CMIP6_StormSurges}, so a surrogate trained on one forcing source may encounter shifted wind and pressure statistics when applied to another. A useful surrogate must therefore preserve station-relevant spatial information, represent flexible temporal dependence, devote sufficient learning capacity to rare but consequential peaks, and remain robust under forcing distribution shift.

To address these requirements, we propose PACT, a \textbf{P}eak-\textbf{A}ware \textbf{C}ross-Attention Graph \textbf{T}ransformer for forcing-driven station-level storm-surge emulation. PACT represents gridded wind and atmospheric pressure fields over the selected spatial domain as graphs and uses a graph encoder to produce spatial node embeddings at each time step. In order to better model how different water stations interact with forcing inputs, PACT introduces a learned station query that performs cross-attention over the graph embeddings, allowing the model to adaptively emphasize the subregions of the atmospheric field most relevant to a particular station and event. A temporal Transformer encoder then models long-range dependence and variable lag structure across the history of station-conditioned tokens. Finally, a horizon-query decoder generates multi-step predictions by allowing each lead time to attend to the shared temporal memory in a horizon-specific way. This design directly targets the spatial selectivity and temporal lag structures that arise when mapping broad atmospheric fields to localized station-level surge response.

PACT also includes a peak-aware learning design to address the imbalance between common weak-surge conditions and rare high-impact events. The model uses an auxiliary peak-aware prediction head in parallel with the main prediction and a tailored objective that augments standard trajectory loss with tail-focused learning on peak-dominated samples. A horizon-wise slope regularizer further encourages coherent multi-step evolution and discourages unrealistic jagged trajectories. As a result, PACT is designed to improve rare-event fidelity without sacrificing the full seasonal water-level trajectory. The same design principles are not specific to storm surge alone. Because PACT combines graph-based representation of spatial fields, query-based extraction of location-relevant information, temporal attention for lagged dynamics, and tail-aware learning, it provides a general template for geophysical surrogate modeling problems in which localized station-level responses must be predicted rapidly from regional gridded environmental forcing.

We evaluate PACT on multiple tide-gauge stations using NCEP/NCAR reanalysis forcing~\cite{ncep} and further assess robustness under forcing distribution shift using five CMIP6 climate-model datasets. Beyond standard metrics such as RMSE and MAE, we use peak-sensitive diagnostics analysis to evaluate whether the model preserves risk-relevant behavior in the tail of the surge distribution. Through in-dataset and cross-dataset experiments, we show that PACT improves both bulk accuracy and peak fidelity while remaining computationally efficient for repeated large-scale surrogate evaluation across various forcing distributions.

\section{Datasets}

\subsection{Atmospheric Forcing Products}

In this work, we use both reanalysis and climate model atmospheric forcing products to drive the hydrodynamic simulations. These products provide the surface wind speeds and atmospheric pressure fields that drive storm surge.

Specifically, we use the NCEP/NCAR Reanalysis dataset~\cite{ncep} over the period 1979--2014 to provide historical forcing. This dataset is a widely used product that combines observations with a numerical weather prediction system to produce a temporally consistent record of atmospheric variables~\cite{rudeva2011,colle2010new}.

We also use forcing products from the Coupled Model Intercomparison Project Phase 6 (CMIP6)~\cite{Eyring2016}, primarily for generalization studies. CMIP6 products are generated by a suite of global climate models that simulate the Earth’s climate under different greenhouse-gas emission scenarios. They provide both historical and future data, making them well-suited for evaluating the robustness of the proposed method under shifts in forcing source and climate state. We select a subset of CMIP6 models that have been widely used in coastal impact studies and that provide the data at suitable spatial and temporal resolutions. The selection is based on both their representativeness of the distribution of historical-to-future changes in mean surface temperature across the broader CMIP6 ensemble and their ability to reproduce historical environmental fields most relevant to tropical cyclones~\cite{begmohammadi2025}. In addition, prior research has shown that ensembles of CMIP6 models that have included the ones utilized in our study resolve ETC frequencies~\cite{priestley2020} to be broadly consistent with reanalysis data, and that the models used in this study perform comparably to similar models in resolving ETC frequencies and intensities~\cite{gore2023}. The selected models and their spatial resolutions are summarized in Table~\ref{table:cmip6}.  The historical simulations span 1979--2014, while the future projections cover 2070--2100 under the Shared Socioeconomic Pathway scenario SSP5-8.5 (SSP5-8.5).

These forcing products differ substantially in both origin and statistical characteristics. Reanalysis combines observations with data assimilation to reconstruct historical atmospheric states, whereas CMIP6 forcings are generated by free-running climate models with model-specific parameterizations and internal variability. This distinction is important because it induces a meaningful distribution shift between reanalysis and climate-model forcings, allowing us to test whether the learned emulator captures transferable forcing--surge relationships rather than overfitting to the statistical idiosyncrasies of a single data source.

\begin{table}[ht]
\centering
\caption{List of model names, horizontal spatial resolutions, and key references. \label{table:cmip6}}
\begin{tabular}{ccccc}
No.                      &  Model Name     & Hor. Resolution (Lon. by Lat. in Degrees)  & Key References    \\ \hline
1                & AWI-CM-1-1-MR  & $1.9^{\circ} \times  1.2^{\circ}$     & \cite{awi}       \\
2 & CNRM-CM6-1  &   $1.4^{\circ} \times  1.4^{\circ}$ &  \cite{cnrm}\\
3 & EC-Earth3  &   $0.7^{\circ} \times  0.7^{\circ}$ &  \cite{earth}\\
4 & MPI-ESM1-2-HR  &   $0.9^{\circ} \times  0.9^{\circ}$ &  \cite{mpi}\\
5 & MRI-ESM2-0  &   $1.1^{\circ} \times  1.1^{\circ}$ & \cite{mri}\\
6 & NCEP-NCAR & $1.9^\circ \times 1.875^\circ$ & \cite{ncep}
\end{tabular}
\end{table}

\subsection{ADCIRC Hydrodynamic Simulations}

In this study, we use the ADCIRC model to simulate surge response. ADCIRC is a popular model within the coastal risk analysis and storm surge prediction communities~\cite{fleming2008, begmohammadi2025}. It employs the finite element method over unstructured triangular meshes to numerically solve the governing equations of shallow water fluid flow~\cite{adcirc} for small-scale free surface displacements. These governing equations are the vertically-integrated turbulent incompressible Reynolds equations, simplified with the hydrostatic pressure and Boussinesq approximations. 

\begin{figure}[ht]
  \centering
  \includegraphics[width=1.0\textwidth]{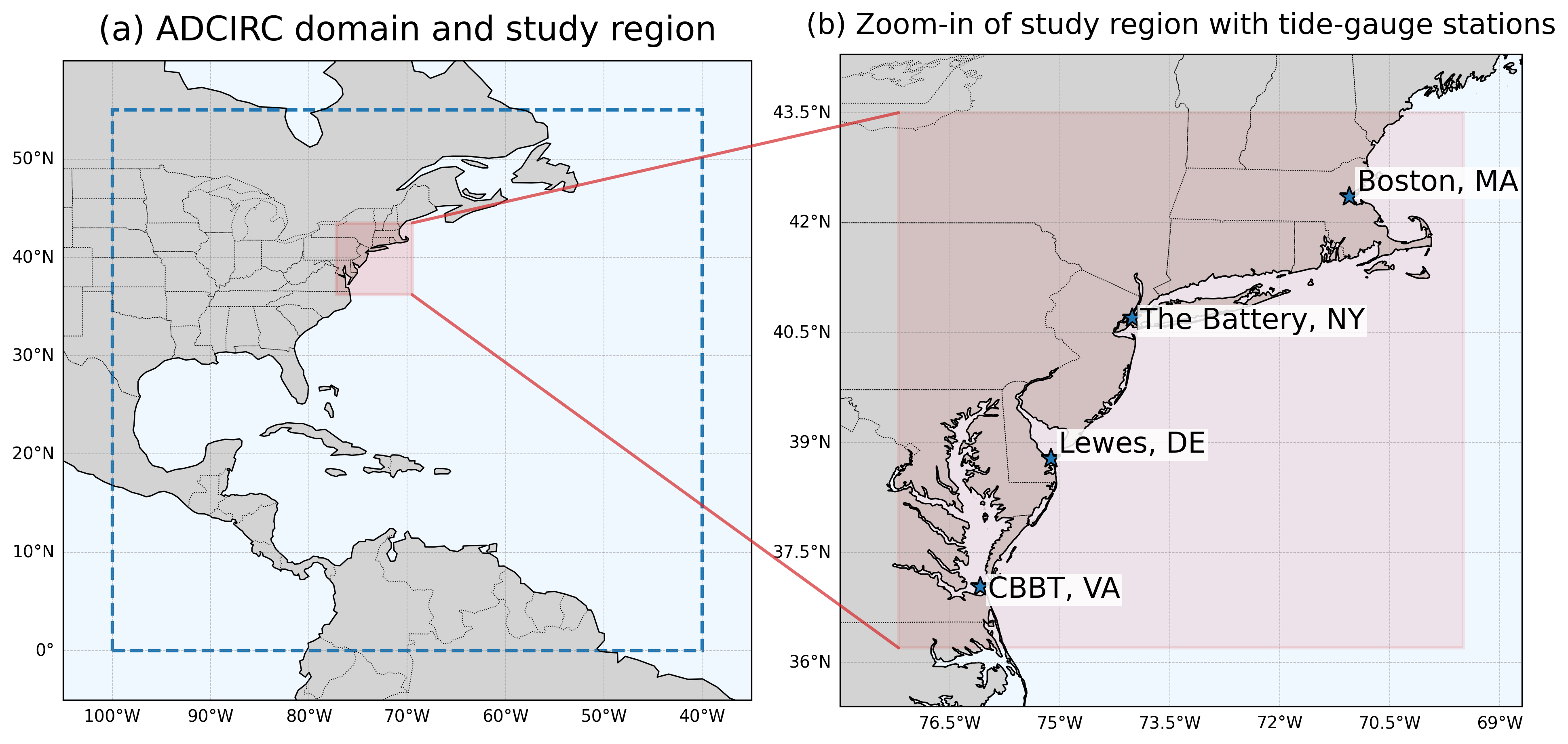}
  \caption{Overview of the computational setup used in this study. (a) The ADCIRC computational domain over the western North Atlantic, with the fixed local study region highlighted in red. The dashed blue line defines the extents of the computational mesh used for numerical simulations. (b) Enlarged view of the study region showing the four tide-gauge stations used in the analysis: Lewes, Delaware; Boston, Massachusetts; Chesapeake Bay Bridge–Tunnel (CBBT), Virginia; and The Battery, New York.\label{fig:domain}}
\end{figure}

ADCIRC is forced during the winter months (November 1 to March 31 of the following year) for both historical (1979-2014) and future (2070-2100) time periods. The simulations are performed on a validated unstructured coastal mesh whose resolution ranges from approximately 1 km in nearshore regions to roughly 100 km in the open ocean~\cite{marsooli2018}. Atmospheric forcing fields are cropped to only fully envelope the ADCIRC mesh, and wind stresses and pressures are applied at evenly distributed nodes throughout the region. In addition to atmospheric forcing, tidal boundary conditions are prescribed using eight tidal constituents from TPXO9-Atlas~\cite{tpxo9}. The simulations output continuous time series of water elevations at every node within the mesh. After the completion of a simulation, time series data at nodes of interest were extracted. As shown in Figure~\ref{fig:domain}(\textbf{b}), we choose four locations throughout the Northeast US: The Chesapeake Bay Bridge Tunnel, VA (CBBT), Lewes, DE, The Battery, NY, and Boston, MA. These locations correspond to existing NOAA tidal gauge stations that have continuously recorded water elevation levels over time, and were selected based on their representation of distinct regions throughout the coast. These sites represent some of the most densely populated coastal regions in the US and are consistent with those that are commonly analyzed by hydrodynamic and risk analysis studies~\cite{marsooli2018, begmohammadi2025, lin2019, ayyad2022}.

For each forcing case, we perform two ADCIRC simulations: one forced by both tides and atmospheric forcing, and the other forced by tides alone. We define the surge response as the residual of the water level produced by these two simulations, and use this surge signal as the target for training and evaluation. Let $\eta_t^{\mathrm{wl}}$ denote the simulated water level forced by tide plus atmospheric forcing, and $\eta_t^{\mathrm{tide}}$ the tide-only water level. We then define the residual storm surge as
\begin{equation}
\eta_t^{\mathrm{surge}} = \eta_t^{\mathrm{wl}} - \eta_t^{\mathrm{tide}}.
\end{equation}

\subsection{Graph Data Generation}
\label{sec:graph_data_generation}

We represent the atmospheric forcing input as a short sequence of graphs defined on the shared fixed region of interest. This yields a consistent supervised-learning setup across forcing products with different native temporal resolutions while preserving the native hourly resolution of the residual surge target.

\subsubsection{Temporal alignment and sample construction.}
The atmospheric forcing and surge target data are available at different temporal resolutions. NCEP forcing is provided every 6 h, CMIP6 forcing every 3 h, and ADCIRC water-level outputs hourly. To define a common learning problem across forcing products, we anchor each sample at an emulation origin time $t$ on a 6 h grid.

Considering the existence of potential delayed surge response to atmospheric forcing, we construct each sample using a short sequence of three forcing graphs at times $t-12\mathrm{h}$, $t-6\mathrm{h}$, and $t$. This allows the model to capture the temporal evolution of the atmospheric state leading to the emulation origin, which is important for accurately predicting the surge response. The target for each sample is defined as the hourly residual surge sequence at the selected station from the emulation origin to the next 5 h, i.e., from time $t$ to $t+5\mathrm{h}$. This setup allows us to evaluate the model's ability to predict both the immediate surge response at the emulation origin and its evolution over a short lead time. Formally, the input is
\begin{equation}
\left( G_{t-12\mathrm{h}},\, G_{t-6\mathrm{h}},\, G_t \right),
\end{equation}
and the corresponding target vector is
\begin{equation}
\mathbf{y}_t =
\left[
\eta^{\mathrm{surge}}_{t},
\eta^{\mathrm{surge}}_{t+1\mathrm{h}},
\eta^{\mathrm{surge}}_{t+2\mathrm{h}},
\eta^{\mathrm{surge}}_{t+3\mathrm{h}},
\eta^{\mathrm{surge}}_{t+4\mathrm{h}},
\eta^{\mathrm{surge}}_{t+5\mathrm{h}}
\right] \in \mathbb{R}^6.
\end{equation}

For NCEP, the required forcing snapshots are available directly on the 6 h grid. For CMIP6, which is available every 3 h, we retain only those forcing fields whose timestamps exactly match the required input times, without temporal interpolation or averaging. The ADCIRC surge outputs are extracted directly from the hourly residual surge time series at the emulation origin and subsequent lead times. Samples near the end of a simulation period are discarded when the full target horizon is unavailable.

\subsubsection{Local forcing graph construction.}
At each aligned time index, the atmospheric forcing field over the shared region of interest is represented as a graph
\begin{equation}
G_t = (\mathcal{V}, \mathcal{E}, \mathbf{X}_t),
\end{equation}
where each node corresponds to one latitude--longitude grid point in the forcing field. The node set $\mathcal{V}$ and edge set $\mathcal{E}$ are fixed within each forcing product, while the node features in $\mathbf{X}_t$ vary with the atmospheric state.

We use a 4-neighbor grid graph in which each node is connected to its immediate horizontal and vertical neighbors on the latitude--longitude grid. The edge set is therefore defined as
\begin{equation}
\mathcal{E} = \{(i,j): j \in \mathcal{N}_4(i)\},
\end{equation}
where $\mathcal{N}_4(i)$ denotes the set of four-neighbor grid points adjacent to node $i$. This construction preserves local spatial adjacency while maintaining a sparse graph representation.

Each node $i \in \mathcal{V}$ is assigned a five-dimensional feature vector
\begin{equation}
\mathbf{x}_{t,i} =
\left[
\mathrm{lat}_i,\,
\mathrm{lon}_i,\,
u_{t,i},\,
v_{t,i},\,
p'_{t,i}
\right] \in \mathbb{R}^5,
\end{equation}
where $\mathrm{lat}_i$ and $\mathrm{lon}_i$ are the spatial coordinates of the grid point, $u_{t,i}$ and $v_{t,i}$ are the horizontal wind components, and $p'_{t,i}$ is the centered pressure anomaly defined in Section~\ref{anomaly}. Each training sample is therefore represented as a short sequence of forcing graphs paired with the corresponding residual surge trajectory. Although the node features are defined on a regular grid, the graph representation maintains permutation equivariance without the need for explicit ordering, and allows for flexible incorporation of irregular spatial structures if needed in future extensions.

\subsection{Pressure-Anomaly Preprocessing via Spatial Mean Centering}
\label{anomaly}
Atmospheric pressure affects storm-surge dynamics primarily through spatial pressure gradients rather than through absolute pressure level. Motivated by this structure, we preprocess the pressure field by removing its instantaneous spatial mean over the shared region of interest at each time step.

Let $p_{t,i}$ denote the raw sea-level pressure at node $i \in \mathcal{V}$ and time $t$. We compute the spatial mean
\begin{equation}
\bar{p}_t = \frac{1}{|\mathcal{V}|} \sum_{i \in \mathcal{V}} p_{t,i},
\end{equation}
and define the centered pressure anomaly
\begin{equation}
p'_{t,i} = p_{t,i} - \bar{p}_t.
\end{equation}
We refer to this transformation as \emph{spatial mean centering of pressure}. By removing the spatial mean, we focus the model's attention on the relevant pressure anomalies and gradients that directly contribute to surge response, rather than on the overall pressure level, which may not be directly related to surge dynamics. This transformation also helps to mitigate potential issues with non-stationarity in the pressure field, as it removes any large-scale trends or biases that may be present in the raw pressure data.

\section{PACT: Peak-Aware Cross-Attention Graph Transformer}

\subsection{Overview of PACT Architecture}
\label{sec:method_overview}

\begin{figure}[ht]
  \centering
  \includegraphics[width=1.0\textwidth]{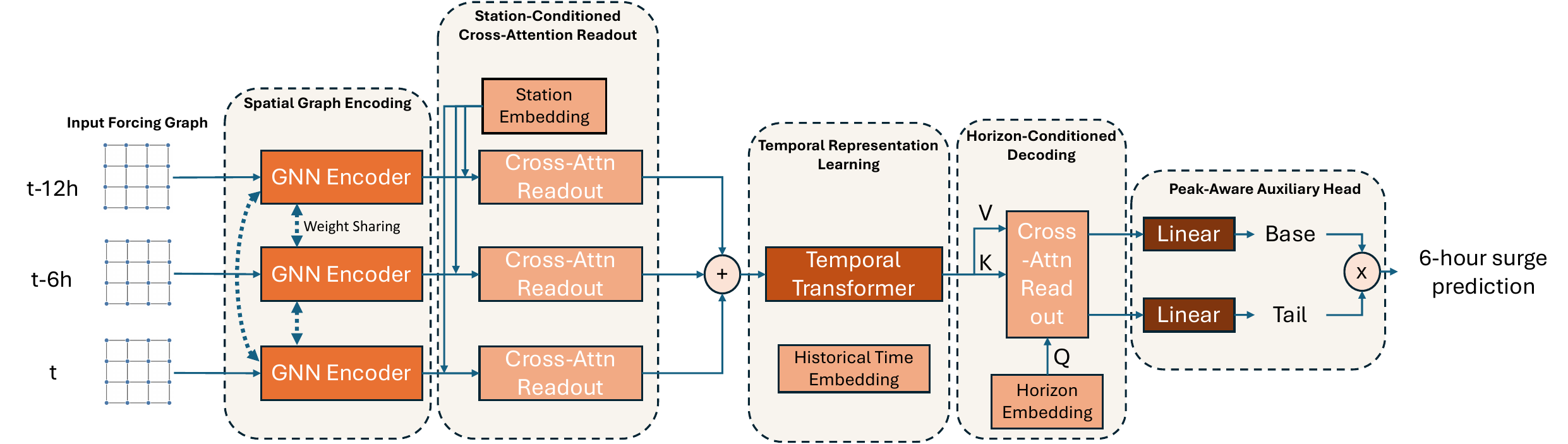}
  \caption{Overall architecture of PACT.\label{fig:arch}}
\end{figure}

PACT predicts short-horizon station-level surge trajectories from gridded atmospheric forcing. Figure~\ref{fig:arch} summarizes the overall data flow. For each emulation origin, the input consists of three forcing graphs defined over a fixed region of interest (ROI) at $t-12\mathrm{h}$, $t-6\mathrm{h}$, and $t$. The graphs share the same spatial support, while their node features vary with the atmospheric forcing at each time. PACT first extracts spatial representations from each forcing graph, then uses a station-conditioned readout to obtain station-relevant forcing tokens. These tokens are further processed by a temporal encoder and a horizon-conditioned decoder to produce multi-step surge predictions. A peak-aware auxiliary branch is included to improve the representation of rare high-surge conditions.

The architecture follows the four main stages. First, a shared GraphSAGE encoder is applied to the three input forcing graphs. This encoder maps node-level atmospheric variables over the fixed ROI into latent spatial embeddings. Because the same encoder is used for all input times, forcing snapshots from $t-12\mathrm{h}$, $t-6\mathrm{h}$, and $t$ are represented in a common latent space, enabling consistent comparison and aggregation across the recent forcing history. Second, PACT converts each encoded forcing graph into a station-conditioned token. For each target station, a learned station representation is used to extract information from the encoded atmospheric field that is most relevant to that station. Applying this readout separately to the three encoded graphs produces a short sequence of station-conditioned tokens, with one token corresponding to each input time. Third, the station-conditioned tokens are augmented with historical time embeddings and passed to a temporal Transformer encoder. This stage models interactions among the recent forcing snapshots and captures how station-relevant atmospheric conditions evolve from $t-12\mathrm{h}$ to $t$. The resulting temporal representation serves as a shared memory for multi-horizon prediction. Fourth, PACT uses horizon-conditioned decoding to generate the short-horizon surge trajectory. Learned horizon representations are assigned to the output lead times and used to extract lead-specific information from the shared temporal memory. The resulting hidden states are mapped to surge predictions, allowing different emulation horizons to emphasize different aspects of the same encoded forcing history. 

In parallel with the main prediction pathway, PACT includes a peak-aware auxiliary branch connected to the learned temporal representation. This branch encourages the model to retain information associated with rare but high-impact surge conditions, improving its ability to represent both ordinary surge evolution and extreme-event behavior. The following subsections describe the spatial graph encoder, station-conditioned readout, temporal Transformer, horizon-conditioned decoder, and peak-aware learning strategy in detail.

\subsection{Spatial Graph Encoding of Local Atmospheric Forcing}

We begin by encoding each station-centered forcing patch as a graph so that the model can learn spatially structured representations of the local atmospheric state before any station-level aggregation is applied. Specifically, given a forcing graph $\mathcal{G}_{t-k}$(where $k=0, 6h, 12h$) at each time step in the input window, we use GraphSAGE inductive framework~\cite{Hamilton_2017_GraphSAGE} to encode its spatial atmospheric structure into a node embedding matrix $\mathbf{U}_{t-k}\in\mathbb{R}^{N\times d}$. This stage learns spatial interactions and local mixing induced by the atmospheric field, such as coherent wind structures and pressure gradients, before any temporal modeling is applied.

Let $\mathbf{X}_{\tau}\in\mathbb{R}^{N\times F}$ denote the node feature matrix at a single input time $\tau \in \{t-12\mathrm{h},\, t-6\mathrm{h},\, t\}$. We initialize node states as $\mathbf{h}^{(0)}_{i}=\mathbf{x}_{\tau,i}$ for node $i\in\{1,\dots,N\}$ and apply a stack of $L$ GraphSAGE layers. At layer $\ell$, each node aggregates information from its neighborhood $\mathcal{N}(i)$ and updates its representation by
\begin{equation}
\mathbf{m}^{(\ell)}_{i} = \mathrm{AGG}\big(\{\mathbf{h}^{(\ell-1)}_{j}: j\in\mathcal{N}(i)\}\big),\qquad
\mathbf{h}^{(\ell)}_{i} = \phi\!\left(\mathbf{W}^{(\ell)}\,[\mathbf{h}^{(\ell-1)}_{i}\ \Vert\ \mathbf{m}^{(\ell)}_{i}] + \mathbf{b}^{(\ell)}\right),
\label{equ:graphsage}
\end{equation}
where $\mathrm{AGG}$ is a permutation-invariant aggregator (mean aggregation in our implementation), $\mathbf{W}^{(\ell)}$ and $\mathbf{b}^{(\ell)}$ are learnable parameters, $\phi$ is a pointwise nonlinearity (LeakyReLU in our implementation), and $\Vert$ denotes concatenation. We apply dropout after each layer during training. The final node embeddings at input time $\tau$ are collected as
\begin{equation}
\mathbf{U}_{\tau}
=
[\mathbf{h}^{(L)}_{1},\dots,\mathbf{h}^{(L)}_{N}]^\top
\in\mathbb{R}^{N\times d}.
\end{equation}
Because the model input consists of three forcing snapshots, we encode each input time independently using shared GraphSAGE weights, yielding the sequence of node-embedding matrices
$\big(\mathbf{U}_{t-12\mathrm{h}},\, \mathbf{U}_{t-6\mathrm{h}},\, \mathbf{U}_{t}\big)$, which capture the spatial structure of the forcing field at each input time.

This per-step spatial encoding is motivated by the need to represent atmospheric forcing in a spatially structured form that is not tied to the ordering of the input data. Storm surge is driven by organized atmospheric patterns, and representing the forcing field as a graph allows information to propagate through explicitly defined spatial adjacencies within the ROI. Unlike tensor-based CNN representations, where neighborhood relationships are determined by array layout, graph message passing is permutation equivariant: the learned representation depends on node features and graph connectivity rather than the arbitrary order in which grid points are stored. This property is useful when working with forcing products that may differ in grid layout, indexing convention, or spatial resolution.

The graph formulation also improves flexibility across datasets. Although the forcing fields used in this study are gridded, reanalysis and climate-model products can have different grid sizes and resolutions. A graph encoder can be applied to these datasets using their corresponding spatial connectivity, without redesigning convolutional kernels, padding rules, or boundary handling that are typically tied to a fixed lattice in CNN-based models. This formulation also provides a natural path to future extensions in which the input forcing or target variables are represented on non-gridded structures, such as irregular atmospheric meshes, station networks, or mesh-based outputs from other numerical models.

We use GraphSAGE as the spatial encoder because it provides a simple inductive message-passing mechanism for local neighborhood aggregation. At each layer, the update in Equation~\ref{equ:graphsage} combines the previous representation of node $i$ with an aggregated message from its neighbors. This allows the encoder to preserve local forcing information while incorporating the surrounding meteorological context. The resulting node-level embeddings retain the spatial structure of the forcing field and are passed to the station-conditioned readout described next.

\begin{figure*}[ht]
  \centering
  \includegraphics[width=0.5\textwidth]{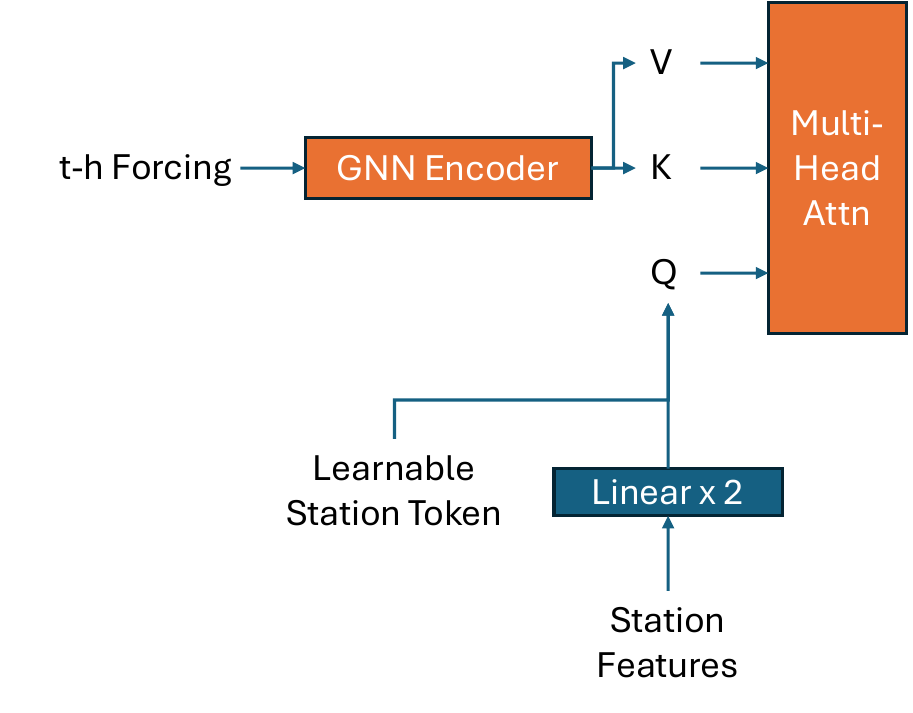}
  \caption{Diagram of station-query readout. The station query is formed by adding a learnable station token to the output of a small station encoder that processes fixed station metadata. The resulting query attends over the node embeddings produced by the GraphSAGE encoder through cross-attention, yielding a station-specific representation of the forcing field.\label{fig:station-query-readout}}
\end{figure*}

\subsection{Station-Conditioned Cross-Attention Readout}
After spatial graph encoding, PACT uses a station-conditioned cross-attention readout to convert the spatially distributed forcing representation into a compact representation tailored to the target tide-gauge station, as illustrated in Figure~\ref{fig:station-query-readout}. This step is important because the surge response at a tide-gauge station is not determined by all forcing nodes equally. Depending on the station location, coastline geometry, and evolving wind and pressure patterns, different parts of the forcing field may be more or less relevant at different times. The readout module therefore allows the model to learn which parts of the local forcing patch are most informative for each station and each input time, while reducing the node-level graph representation to a station-level token that can be passed to the temporal prediction module.

Formally, let $\mathbf{U}_{\tau} \in \mathbb{R}^{N \times d}$ denote the matrix of node embeddings produced by the spatial graph encoder at input time $\tau \in \{t-12\mathrm{h},\, t-6\mathrm{h},\, t\}$, where $N$ is the number of graph nodes in the ROI and $d$ is the latent dimension. The station-conditioned readout uses a station-specific query to attend over these node embeddings. Let $\mathbf{Q}_{s} \in \mathbb{R}^{1 \times d}$ denote the query for station $s$, and let the encoded node embeddings serve as both keys and values,
\begin{equation}
\mathbf{K}_{\tau} = \mathbf{V}_{\tau} = \mathbf{U}_{\tau}.
\end{equation}
The cross-attention readout is then defined as
\begin{equation}
\mathbf{z}_{s,\tau}
=
\mathrm{Attn}(\mathbf{Q}_{s},\mathbf{K}_{\tau},\mathbf{V}_{\tau})
=
\operatorname{softmax}\!\left(
\frac{\mathbf{Q}_{s}\mathbf{K}_{\tau}^{\top}}{\sqrt{d}}
\right)\mathbf{V}_{\tau}
\in \mathbb{R}^{1 \times d}.
\label{equ:station_attention}
\end{equation}

Dropping the singleton dimension, $\mathbf{z}_{s,\tau} \in \mathbb{R}^{d}$ denotes the station-conditioned token for station $s$ at input time $\tau$. In practice, we use multi-head attention~\cite{Vaswani_2017_Attention}, which allows the model to learn multiple complementary spatial relevance patterns over the forcing field and combine them into a single station-level representation~\cite{NEURIPS2019_MultiHead,voita-etal-2019-analyzing}.

To make the readout station specific while retaining one shared model across all stations, we condition the query on fixed station metadata obtained from NOAA~\cite{NOAA_COOPS_WaterLevels}, including latitude, longitude, and meteorological-site elevation. These variables provide simple but important information about the physical setting of each gauge. Latitude and longitude identify where the station lies within the coastal domain and therefore help distinguish stations with different exposure to regional wind and pressure patterns. Elevation provides an additional descriptor of the local observing site. Although these metadata do not fully describe coastline geometry, bathymetry, or local hydraulic connectivity, they give the model a station-dependent reference point that helps it interpret the same atmospheric pattern differently at different gauges.

Let $\mathbf{s} \in \mathbb{R}^{d_s}$ denote the metadata vector for station $s$, and let $\psi(\cdot)$ be a small station encoder implemented as a multilayer perceptron. We define the effective station query as
\begin{equation}
\tilde{\mathbf{q}}_{s}
=
\mathbf{q}_{\mathrm{base}} + \psi(\mathbf{s}),
\qquad
\mathbf{Q}_{s}
=
\tilde{\mathbf{q}}_{s}^{\top}
\in \mathbb{R}^{1 \times d},
\label{equ:station_query}
\end{equation}
where $\mathbf{q}_{\mathrm{base}} \in \mathbb{R}^{d}$ is a learned base query. In this design, the base query captures a general strategy for extracting station-relevant forcing information, while the metadata-dependent term adjusts the query for each specific station. This allows PACT to use one shared graph encoder and cross-attention readout module across stations, while still allowing the spatial aggregation pattern to vary from station to station.

This readout mechanism is well-suited to storm-surge emulation because tide-gauge response is spatially selective. At some emulation times, nearby wind or pressure patterns may dominate the station response; at other times, offshore forcing structures or broader pressure anomalies may be more informative. The station-conditioned query allows the model to assign different attention weights to different nodes in the ROI, producing a time-dependent relevance map over the encoded forcing field. More broadly, this design follows the paradigm of using learned queries to distill relevant information from large token sets~\cite{perceiver,jaegle2022perceiverIO}.

Equation~\ref{equ:station_attention} is applied independently to each of the three encoded forcing graphs. That is, using the same station query $\mathbf{Q}_{s}$, PACT computes
\begin{equation}
\mathbf{z}_{s,t-12\mathrm{h}}
=
\mathrm{Attn}(\mathbf{Q}_{s},\mathbf{U}_{t-12\mathrm{h}},\mathbf{U}_{t-12\mathrm{h}}),
\end{equation}
\begin{equation}
\mathbf{z}_{s,t-6\mathrm{h}}
=
\mathrm{Attn}(\mathbf{Q}_{s},\mathbf{U}_{t-6\mathrm{h}},\mathbf{U}_{t-6\mathrm{h}}),
\end{equation}
and
\begin{equation}
\mathbf{z}_{s,t}
=
\mathrm{Attn}(\mathbf{Q}_{s},\mathbf{U}_{t},\mathbf{U}_{t}).
\end{equation}
These three station-conditioned tokens form the compact temporal input sequence
\begin{equation}
\left(
\mathbf{z}_{s,t-12\mathrm{h}},\
\mathbf{z}_{s,t-6\mathrm{h}},\
\mathbf{z}_{s,t}
\right),
\end{equation}
which summarizes the recent atmospheric forcing history from the perspective of station $s$. This sequence retains adaptive spatial relevance while reducing the encoded forcing fields to three station-level tokens, which are then passed to the temporal representation module described next.

\subsection{Temporal Representation Learning and Horizon-Conditioned Decoding}

Given the sequence of station-conditioned tokens, PACT next models temporal dependence across the recent forcing history in order to capture event-dependent lag structure in surge response. For the three forcing snapshots in the input sequence, the station-query readout produces the corresponding station tokens
$\mathbf{z}_{s, t-12\mathrm{h}}, \mathbf{z}_{s,t-6\mathrm{h}}, \mathbf{z}_{s,t}\in\mathbb{R}^{d}$. Collecting them in temporal order yields
\begin{equation}
\mathbf{Z}
=
\begin{bmatrix}
\mathbf{z}_{s,t-12\mathrm{h}}\\
\mathbf{z}_{s,t-6\mathrm{h}}\\
\mathbf{z}_{s,t}
\end{bmatrix}
\in\mathbb{R}^{3\times d}.
\end{equation}
To encode temporal order, we add learned temporal embeddings for the three input times:
\begin{equation}
\tilde{\mathbf{z}}_{s, \tau} = \mathbf{z}_{s, \tau} + \mathbf{p}_{\tau},
\qquad
\tau \in \{t-12\mathrm{h},\, t-6\mathrm{h},\, t\},
\end{equation}
where $\mathbf{p}_{\tau}\in\mathbb{R}^{d}$ is the learned temporal embedding associated with input time $\tau$. The resulting temporally encoded token sequence is denoted by
\begin{equation}
\tilde{\mathbf{Z}}
=
\begin{bmatrix}
\tilde{\mathbf{z}}_{s,t-12\mathrm{h}}\\
\tilde{\mathbf{z}}_{s,t-6\mathrm{h}}\\
\tilde{\mathbf{z}}_{s,t}
\end{bmatrix}
\in\mathbb{R}^{3\times d}.
\end{equation}

We feed $\tilde{\mathbf{Z}}$ into a temporal Transformer encoder to obtain contextualized representations of the recent forcing history. Specifically, we use the standard multi-head self-attention architecture~\cite{Vaswani_2017_Attention} over the station-token sequence, followed by a position-wise feedforward network, with residual connections and layer normalization. The temporal encoder maps the input sequence into contextualized memory tokens
\begin{equation}
\mathbf{H}
=
\begin{bmatrix}
\mathbf{h}_{s,t-12\mathrm{h}}\\
\mathbf{h}_{s,t-6\mathrm{h}}\\
\mathbf{h}_{s,t}
\end{bmatrix}
\in\mathbb{R}^{3\times d},
\end{equation}
which summarizes the station-specific forcing history in a temporally aware latent space.

Self-attention enables interactions among all input times, so the temporal representation is not constrained to a fixed lag structure and can capture richer dependencies across the forcing history. This flexibility is important for storm surge, where relevant temporal effects may vary across events and can include sustained forcing buildup, abrupt shifts in forcing regime, and delayed response to evolving atmospheric conditions. The temporal encoder produces a shared latent memory that summarizes the recent forcing history and its event-dependent temporal interactions. From this shared representation, PACT then generates horizon-specific predictions through a query-based decoder that assigns a dedicated readout to each output horizon, allowing different lead times to attend to different aspects of the encoded history.

Starting with the contextualized memory tokens $\mathbf{H}$ output by the temporal Transformer, we introduce a collection of learned horizon queries $\{\mathbf{q}_h\}_{h=0}^{5}$ with $\mathbf{q}_h \in \mathbb{R}^{d}$, where $h$ indexes the output horizon corresponding to prediction times $t, t+1\mathrm{h}, \dots, t+5\mathrm{h}$. For each horizon, we obtain a horizon-specific context vector by cross-attending to the shared temporal memory:
\begin{equation}
\mathbf{c}_h
=
\mathrm{Attn}(\mathbf{q}_h, \mathbf{H}, \mathbf{H})
\in \mathbb{R}^{d}.
\end{equation}
Stacking the resulting context vectors yields
\begin{equation}
\mathbf{C}
=
\begin{bmatrix}
\mathbf{c}_{0}\\
\mathbf{c}_{1}\\
\vdots\\
\mathbf{c}_{5}
\end{bmatrix}
\in \mathbb{R}^{6 \times d},
\end{equation}
which is then mapped by the output head to the corresponding 6-step surge predictions.

This horizon-query decoder avoids forcing all output times to share a single pooled temporal representation. In storm-surge prediction, the informative content of recent forcing history is horizon dependent: the prediction at the emulation origin may rely more heavily on the most recent atmospheric state, whereas later hourly predictions may benefit more from earlier forcing snapshots and their temporal interactions. Assigning a dedicated query to each output horizon therefore allows the model to form horizon-specific cross-attention weights over the shared memory $\mathbf{H}$ and extract distinct context vectors without requiring separate encoders for different lead times. This design follows the broader paradigm of query-based decoders for structured outputs, in which learned query vectors index different output slots and retrieve relevant information through cross-attention~\cite{jaegle2022perceiverIO}. The resulting context vectors $\mathbf{C}$ are then passed to the prediction head to generate the final multi-horizon surge predictions. Although this architecture provides a flexible mechanism for general multi-horizon prediction, rare surge extremes remain difficult and motivate the peak-aware design introduced next.

\subsection{Peak-Aware Designs}
While the preceding components provide adaptive spatial aggregation, temporal memory modeling, and horizon-conditioned decoding, accurate storm-surge emulation also requires sufficient sensitivity to rare extreme events. In practice, standard multi-horizon regression is dominated by the much more frequent non-extreme samples, which can lead to underestimation of peak magnitude and reduced fidelity in the upper tail. To address this issue, PACT incorporates a peak-aware design that combines an auxiliary prediction pathway with an extreme-sensitive training objective. The auxiliary head provides dedicated modeling capacity for peak-focused prediction, while the loss design strengthens the optimization signal on peak-dominated samples and encourages temporally coherent emulation.

\subsubsection{Peak-Aware Auxiliary Head}
We first introduce a peak-aware auxiliary prediction head to provide dedicated representational capacity for rare extreme events that may be underemphasized by the main emulation pathway. Two heads are used to allow the model to allocate dedicated capacity to learning extreme corrections without compromising the base performance on typical conditions, which can be overwhelmed by the dominant average-error objective in a single-head design.

For each emulation horizon $h$, the decoder produces a horizon context vector $\mathbf{c}_h\in\mathbb{R}^{d}$. The dual prediction heads output a base prediction $\hat y^{\text{base}}_h$ that serves as the primary surge estimate and a tail residual $r^{\text{tail}}_h$ intended to correct the base prediction in extreme regimes:
\begin{equation}
\hat y^{\text{base}}_h = f_{\text{base}}(\mathbf{c}_h),\qquad
r^{\text{tail}}_h = f_{\text{tail}}(\mathbf{c}_h),
\end{equation}
where $f_{\text{base}}(\cdot)$ and $f_{\text{tail}}(\cdot)$ are MLPs. For numerical stability, we apply a bounded nonlinearity to the tail residual,
\begin{equation}
\tilde r^{\text{tail}}_h = c\,\tanh\!\left(\frac{r^{\text{tail}}_h}{c}\right),
\end{equation}
with clip parameter $c>0$. We combine the two branches through a gated residual formulation:
\begin{equation}
\hat y_h
=
\hat y^{\text{base}}_h
+
g \cdot \alpha \cdot \tilde r^{\text{tail}}_h,
\end{equation}
where $\alpha\in(0,1)$ is a learned global scaling parameter (implemented via a sigmoid) and $g\in(0,1)$ controls the contribution of the tail residual. Because both heads are conditioned on the same horizon representation, this formulation should be viewed as a soft gated-residual parameterization rather than a hard decomposition of typical and extreme dynamics; the effective tail contribution is therefore the gated residual $g \cdot \alpha \cdot \tilde r^{\text{tail}}_h$, not the raw residual $\tilde r^{\text{tail}}_h$ alone.

In our implementation, we use a window gating mechanism: a single gate value is predicted per sample (i.e., per input window) from a pooled summary of the horizon context vectors $\{\mathbf{c}_h\}_{h=0}^{5}$. Specifically, we form
\begin{equation}
\bar{\mathbf{c}}=\frac{1}{6}\sum_{h=0}^{5}\mathbf{c}_h \in \mathbb{R}^{d},
\end{equation}
and compute the gate as $g=\sigma\!\left(f_{\text{gate}}(\bar{\mathbf{c}})\right)\in(0,1)$, where $f_{\text{gate}}(\cdot)$ is an MLP and $\sigma(\cdot)$ is the sigmoid function. This yields one scalar $g$ for each sample in the batch, which is then shared across all six lead times, encouraging a coherent sample-level assessment of severity and preventing the tail correction from fluctuating independently across horizons. This design allows the model to learn a global gating strategy that activates the tail residual when the overall context indicates a potentially extreme event, while keeping it dormant during typical conditions to preserve the base predictor's performance. By conditioning the gate on a pooled summary of the horizon contexts, the model can integrate information across lead times to make a holistic judgment about event severity, which is crucial for effectively leveraging the tail residual without introducing noise into non-extreme predictions.

We emphasize that the dual-head formulation is not intended to impose a hard decomposition between typical and extreme surge dynamics. Since both heads are conditioned on the same horizon representation $\mathbf{c}_h$, their raw outputs should not be interpreted as individually identifiable physical components. Instead, the design is a soft gated-residual parameterization: $f_{\mathrm{base}}$ provides the primary prediction, while $f_{\mathrm{tail}}$ provides an additional residual correction whose effective contribution is controlled by the learned gate $g$. Therefore, the relevant tail adjustment is the gated residual $g \cdot \alpha \cdot \tilde r^{\text{tail}}_h$, rather than the raw tail-head output alone. When $g$ is small, the final prediction remains close to the base estimate; when $g$ is larger, the residual branch can modify the prediction more strongly for samples that require additional correction.

In summary, the peak-aware prediction head is motivated by the strong imbalance between typical and extreme conditions. Extreme surges are rare, yet they can arise from distinct regimes such as compound forcing and nonlinear coastal responses. A single predictor trained primarily on average-error objectives often allocates insufficient capacity to such events, leading to systematic underestimation of peak values. The proposed two-branch design addresses this issue through specialization: the base branch learns the dominant bulk behavior, while the tail branch focuses on extreme corrections. The gate limits the influence of the tail residual outside the extreme regime, thereby preserving base prediction accuracy while improving peak fidelity. Conceptually, this mechanism is related to mixture-of-experts-style specialization~\cite{shazeer2017, Royer2023RevisitingSM} and residual adaptation, but is implemented here in a lightweight and stable form suitable for multi-horizon emulation. Thus, the auxiliary head preserves the overall input--output formulation of the model while supplying additional representational capacity for peak-focused prediction, which is then reinforced by the extreme-aware training objective described next.

\subsubsection{Peak-Aware Loss Design}

While the auxiliary head provides dedicated capacity for extreme-event modeling, improved peak fidelity also depends on a loss objective that prioritizes rare peak-dominated samples while encouraging physically reasonable short-term evolution. To further encourage accurate predictions on extreme surge events, we use a peak-aware objective that augments a standard mean squared error (MSE) term with a tail-focused component and a horizon-wise slope regularizer. All loss terms are computed in physical units (meters), which keeps hyperparameters interpretable and avoids coupling the objective to any particular target normalization used during optimization. Specifically, the tail component applies an additional MSE term to peak-dominated samples (top-$\rho$ by the ground-truth peak over the whole dataset), increasing gradient signal on rare but high-impact events without overwhelming the bulk objective. We also introduce a slope-matching term that penalizes mismatches in the first derivative across lead times, encouraging physically plausible temporal evolution and reducing jagged multi-horizon trajectories that can arise even when pointwise errors are moderate. 

Let $\hat{\mathbf{y}}_b,\mathbf{y}_b\in\mathbb{R}^{6}$ denote the 6-step prediction and target for sample $b$, where $\mathbf{y}_b=[y_{b,0},\dots,y_{b,5}]^\top$ and $\hat{\mathbf{y}}_b=[\hat y_{b,0},\dots,\hat y_{b,5}]^\top$. With batch size $B$, the base objective is mean squared error:
\begin{equation}
\mathcal{L}_{\text{mse}}
=
\frac{1}{6B}
\sum_{b=1}^{B}
\sum_{h=0}^{5}
\left(\hat y_{b,h}-y_{b,h}\right)^2 .
\end{equation}

Because storm-surge observations are heavy-tailed, an unweighted standard mean squared error (MSE) can underemphasize the prediction of surge levels during extreme events, which are critical for risk assessment but can be overwhelmed by the dominant signal under typical conditions. To mitigate this issue, we therefore augment the base objective with a tail-focused term defined over peak-dominated samples. For each sample, we define the ground-truth peak score over the 6-step horizon as
\begin{equation}
p_b = \max_{h\in\{0,\dots,5\}} (y_{b,h})
\end{equation}
Let $\rho\in(0,1)$ denote the tail fraction (e.g., $\rho=0.05$). Following the training procedure, we compute a peak threshold $\tau_{\text{tail}}$ from the training set as the $(1-\rho)$-quantile of $\{p_b\}$:
\begin{equation}
\tau_{\text{tail}} = \mathrm{Quantile}_{1-\rho}(\{p_b\}_{b\in\text{train}}),
\end{equation}
and define the tail subset $\mathcal{B}_{\text{tail}}=\{b: p_b\ge \tau_{\text{tail}}\}$. The tail loss applies the same standard MSE loss, but averaged only over $\mathcal{B}_{\text{tail}}$:
\begin{equation}
\mathcal{L}_{\text{tail}}
=
\frac{1}{\left|\mathcal{B}_{\text{tail}}\right|}
\sum_{b\in\mathcal{B}_{\text{tail}}}
\frac{1}{6}\sum_{h=0}^{5}
\big(\hat y_{b,h}-y_{b,h}\big)^2.
\end{equation}
The combined objective is
\begin{equation}
\mathcal{L}
=
\mathcal{L}_{\text{mse}} + \lambda_{\text{tail}}\mathcal{L}_{\text{tail}},
\end{equation}
where $\lambda_{\text{tail}}\ge 0$ controls the additional emphasis on peak-dominated samples. We compute $\tau_{\text{tail}}$ once from the training distribution and keep it fixed during training, which makes the tail subset definition stable and avoids batch-dependent threshold fluctuations.

As we set up the problem as a regression task for every 6-hour surge level instead of using an autoregressive manner, the model can occasionally produce horizon trajectories that are unrealistically jagged, even when pointwise errors are moderate. To encourage physically plausible temporal evolution across lead times, we introduce a slope-matching term that aligns first derivatives along the horizon axis. For $h=1,\dots,5$, we define the predicted and ground-truth slopes as
\begin{equation}
\Delta\hat y_{b,h}=\hat y_{b,h}-\hat y_{b,h-1},\qquad
\Delta y_{b,h}=y_{b,h}-y_{b,h-1},
\end{equation}
and slope errors $$e_{b,h}=\Delta\hat y_{b,h}-\Delta y_{b,h}$$. We penalize these errors using a robust Charbonnier loss function~\cite{Barron_2019_CVPR} $\rho(x) = \sqrt{x^2 + \epsilon^2}$ where $\epsilon=0.001$ is a small constant. The slope loss is then defined as the average robust Charbonnier loss across all samples and slopes:
\begin{equation}
\mathcal{L}_{\text{slope}}
=
\frac{1}{B}\sum_{b=1}^{B}
\frac{1}{5}\sum_{h=1}^{5}\rho(e_{b,h}).
\end{equation}
The final peak-aware training objective is
\begin{equation}
\mathcal{L}_{PeakAware}
=
\mathcal{L}_{\text{mse}} + \lambda_{\text{tail}}\mathcal{L}_{\text{tail}} + \lambda_{\text{slope}}\mathcal{L}_{\text{slope}},
\end{equation}
where $\lambda_{\text{slope}}\ge 0$ controls the strength of the trajectory regularizer.

In evaluation, we report standard RMSE/MAE computed over all samples and horizons, and we additionally compute peak-centric metrics restricted to peak-dominated subsets defined by the same peak score $p_b$ (e.g., top-$\rho$ by ground-truth peak). This directly measures performance on emulation of surge levels during extreme events, which are critical for risk assessment but can be under-emphasized by average-error metrics. By combining the dual-head architecture with this peak-aware loss, we aim to improve the model's peak prediction accuracy without sacrificing overall performance. The tail loss provides a stronger gradient signal for extreme events, while the slope regularizer encourages physically plausible multi-horizon trajectories, both of which are important for accurate and reliable storm-surge emulation in high-impact regimes. Together with the auxiliary head, this objective forms the complete peak-aware design of PACT, enabling the model to better capture both bulk surge evolution and rare extreme behavior.

\section{Experimental Protocol}
\label{sec:experiment}

We evaluate PACT under a unified experimental protocol designed to compare it against a strong graph-based baseline under consistent split and optimization settings. This section describes the baseline model, the dataset split settings, and the training configuration used throughout the study.

\subsection{Baseline Model}

We compare PACT with two graph-based baselines of increasing complexity. The first is a simple GNN baseline that uses GraphSAGE to encode only the atmospheric forcing field at the current time step, without incorporating any historical forcing information. The second is a spatio-temporal GNN baseline, denoted ST-GNN, which uses the same historical forcing inputs as PACT. In ST-GNN, GraphSAGE is first applied to each forcing snapshot to extract spatial node-level features, followed by global pooling to produce a fixed-length station-level representation at each input time. The resulting sequence of pooled representations is then passed to an LSTM to model temporal dependencies. An architecture diagram of this ST-GNN baseline is provided in the Appendix.

For experiments without historical forcing, the model uses only the contemporaneous forcing input and no temporal encoder is required. This baseline reflects a conventional design in which spatial information is compressed through fixed global aggregation before prediction, and thus provides a direct reference for evaluating the advantages of PACT’s adaptive station-specific aggregation, attention-based temporal modeling, and peak-aware design.

\subsection{Dataset Split Settings}\label{sec4.2}

To assess both within-dataset generalization and robustness under forcing shift, we consider the following dataset split settings.

\begin{itemize}
    \item \textbf{NCEP reanalysis.} For the NCEP dataset, we use 1979--2000 (22 years) for training, 2001--2007 (7 years) for validation, and 2008--2014 (7 years) for testing. This split is applied consistently across all four stations. We refer to this setting as \textbf{Past-Only}.
    
    \item \textbf{CMIP6 climate-model forcings.} For each of the five CMIP6 datasets, we consider two within-dataset evaluation settings:
    \begin{itemize}
        \item \textbf{Future-Period:} 1979--2008 (30 years) for training, 2009--2014 (6 years) for validation, and 2070--2099 (30 years) for testing. This setting is designed to assess generalization from the historical period to future climate conditions.
        
        \item \textbf{Past-Only:} 1979--2000 (22 years) for training, 2001--2007 (7 years) for validation, and 2008--2014 (7 years) for testing. This setting mirrors the split used for the NCEP dataset and evaluates performance within the historical period.
    \end{itemize}
\end{itemize}

For the CMIP6 datasets, we train and evaluate models only on the Battery station in order to focus the climate-forcing analysis on cross-forcing robustness rather than multi-station variation.

In addition to these within-dataset settings, we also perform cross-dataset transfer experiments in which a model trained on one forcing source is evaluated on a different forcing source. In this transfer setting, we report results under three testing periods: \textbf{Past-Only}, \textbf{Future-Period}, and \textbf{All-Year}. Here, \textbf{All-Year} denotes an evaluation-only setting in which predictions are assessed over the entire available test period of the target dataset, including both past years and future years.

\subsection{Training Configuration}

Unless otherwise noted, all models are trained separately for each station--dataset pair; that is, no training run mixes samples from different stations or different forcing datasets. All training experiments use a batch size of 256, an initial learning rate of 0.005, and 300 training epochs. Optimization is performed using Adam with a weight decay of \(10^{-5}\). The learning-rate schedule consists of a linear warmup over the first 5 epochs followed by cosine decay for the remaining epochs, with a minimum learning rate of \(10^{-6}\).

Model training is performed on four NVIDIA H100 GPUs with an Intel Xeon Platinum 8468 CPU host. The ADCIRC simulations used to generate the surge targets are parallelized on the Princeton Della cluster using AMD EPYC 9654 and Intel Cascade Lake processors, with most simulations using approximately 64--128 CPU cores depending on the dataset and year. These computational settings provide the simulation data used for training and evaluation, while the learned models themselves remain lightweight enough for rapid inference once trained.

Under this protocol, the following sections evaluate PACT in terms of predictive accuracy, peak-event fidelity, robustness across forcing datasets, and computational efficiency.

\section{Results}
\label{sec:results}

In this section, we report results on the NCEP reanalysis dataset under the \textit{Past-Only} setting. We first compare the overall emulation performance of PACT and the baseline model across the four stations, then evaluate their behavior on peak surge values, which are of particular importance for coastal hazard assessment. Finally, we assess computational runtime to quantify the practical efficiency of the proposed approach. Taken together, these results characterize PACT from three complementary perspectives: average predictive accuracy, fidelity on rare extreme events, and inference efficiency.

\subsection{Overall Predictive Performance Across Stations}

\begin{figure}[ht]
  \centering
  \includegraphics[width=1.0\textwidth]{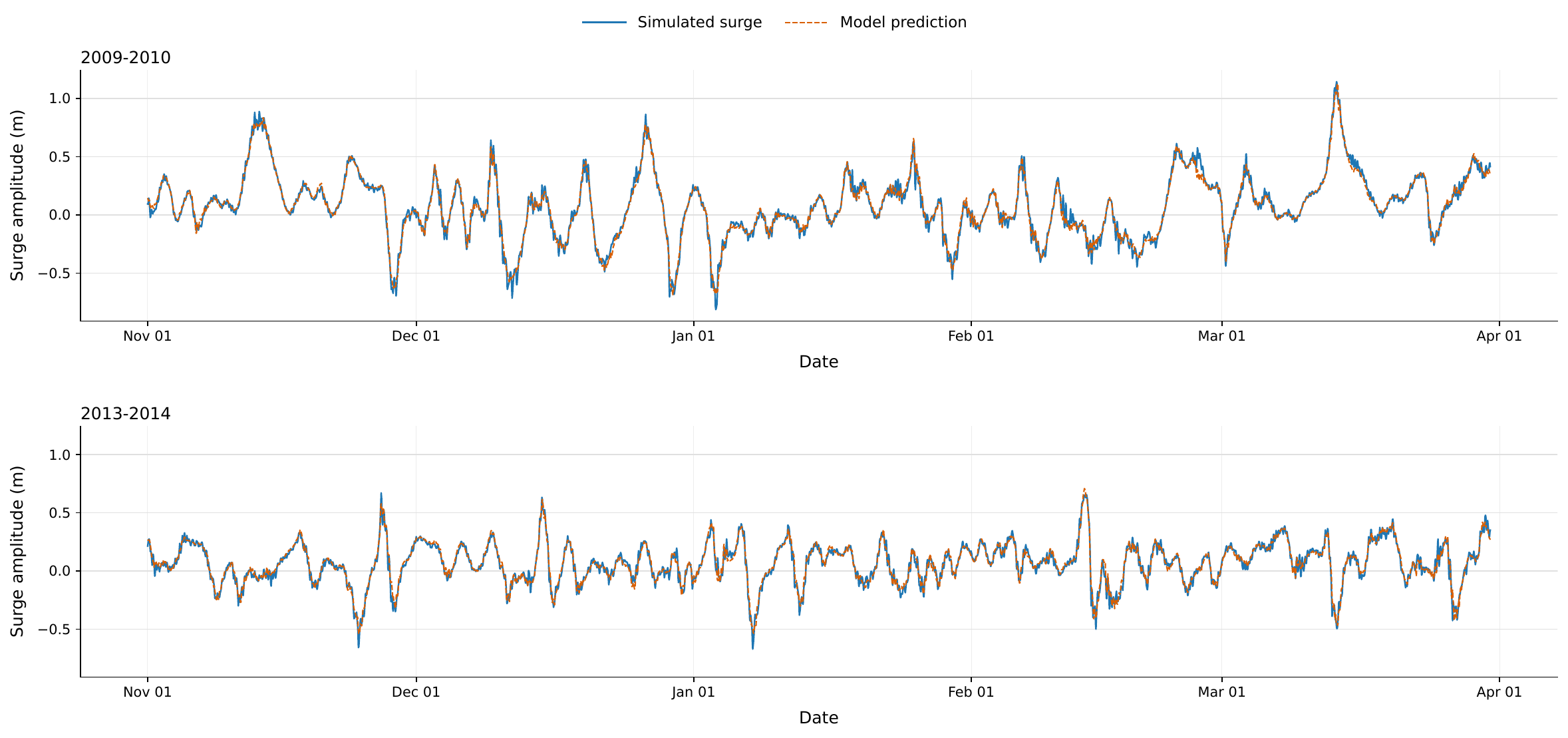}
  \caption{Qualitative examples of surge level time series predictions at Battery, NY, made by PACT with peak aware loss on NCEP Reanalysis dataset. \label{fig:qual-results}}
\end{figure}

To assess the overall predictive capability of the proposed model, we compare PACT with the baseline network across the four stations under the NCEP \textit{Past-Only} setting. Table~\ref{table:overall} summarizes the overall emulation accuracy on the NCEP Reanalysis dataset across the four stations, reported in terms of RMSE and MAE in meters. As described in Section~\ref{sec:experiment}, models are trained separately for each station using only the corresponding station's NCEP Reanalysis data, with no mixing of samples across stations or datasets. As shown in Figure~\ref{fig:qual-results}, PACT with the peak-aware loss closely follows the ADCIRC-simulated hourly surge trajectories over the full season, including both short-term fluctuations and major peak events.

\begin{table}[ht]
\centering
\caption{Overall performance on the NCEP Reanalysis dataset at four NOAA tide-gauge stations (Battery, Boston, Lewes, and CBBT). Models are trained and evaluated independently per station. Simple GNN baseline does not use historical forcing inputs while other three models consider 12 hours historical forcing input.\label{table:overall}}
\scalebox{0.6}{
\begin{tabular}{cccccccccc}
\multirow{2}{*}{Methods}     & \multirow{2}{*}{Loss Function}                                        & \multicolumn{2}{c}{Battery Error}             & \multicolumn{2}{c}{Boston Error}              & \multicolumn{2}{c}{Lewes Error}               & \multicolumn{2}{c}{CBBT Error}                \\ \cline{3-9} 
                                                                    & & RMSE                  & MAE                   & RMSE                  & MAE                   & RMSE                  & MAE                   & RMSE                  & MAE                   \\ \hline
Simple GNN Baseline &       Mean Squared Error                                & 0.0659                & 0.0480                & 0.0495                & 0.0354                & 0.0655                & 0.0511                & 0.0631                & 0.0473                \\
ST-GNN Baseline    &     Mean Squared Error                              & 0.0536                & 0.0384                & 0.0426                & 0.0310                & 0.0447                & 0.0341                & 0.0469                & 0.0348                \\
PACT(Ours)     &  Mean Squared Error                & 0.0349                & 0.0253                & 0.0282                & 0.0209                & 0.0279                & 0.0210                & 0.0309               & 0.0237                \\
PACT(Ours) + Peak Aware Loss &  Peak Aware Loss & \textbf{0.0337} & \textbf{0.0246} & \textbf{0.0274} & \textbf{0.0203} & \textbf{0.0276} & \textbf{0.0208} & \textbf{0.0306} & \textbf{0.0235}
\end{tabular}
}
\end{table}

Comparing ST-GNN against simple GNN baselines, the results show that incorporating a 12-hour forcing history already benefits the baseline model. With the same 12-hour forcing input, the proposed PACT model consistently achieves substantially lower errors than the baseline across all stations. Compared with the ST-GNN baseline, PACT with the standard MSE loss reduces RMSE by 0.0144--0.0187\,m (34--38\%) and MAE by 0.0101--0.0131\,m (32--38\%). Furthermore, replacing the standard MSE objective with the proposed peak-aware training objective yields additional consistent improvements in overall accuracy, with up to 0.0012\,m reduction in RMSE and 0.0007\,m reduction in MAE. For simplicity in the following discussion, we refer to PACT trained with the standard MSE loss as the \textbf{base PACT configuration}, and PACT trained with the peak-aware loss as the \textbf{best PACT configuration}.

To keep the main text focused, the detailed ablation studies on spatial mean centering of pressure anomalies and on the peak-aware design are presented in~\ref{appendix: ablation-spatial-mean} and~\ref{appendix:ablation-loss}, respectively. These additional results are consistent with the main comparisons reported here and further clarify the contribution of each component to the final performance of PACT. Having established the overall accuracy gains of PACT, we next examine whether these improvements extend to the more demanding task of peak surge prediction.

\subsection{Performance on Peak Surge Events}

Although overall error metrics are informative, they do not fully characterize model behavior on rare extremes. We therefore next examine performance on peak surge values using peak-focused metrics that more directly assess the prediction of high-magnitude surge events. To avoid redundancy, the main manuscript reports only the top 5\% peak-evaluation results for the NCEP Reanalysis dataset across the four stations, as summarized in Table~\ref{table:5-peak}, while the corresponding top 1\% and top 10\% threshold results are provided in~\ref{appendix:additional-peak-thresholds}. Each station is trained and evaluated independently.

For each station, peak events are identified using thresholds defined from the observed surge distribution. We report four peak-focused metrics: Peak RMSE, Peak MAE, Peak Mean Signed Error, and Peak Max Absolute Error. As expected, errors are higher for these peak-focused evaluations than for the overall RMSE and MAE reported earlier, indicating that extreme surge prediction is more challenging than bulk emulation. From Table~\ref{table:5-peak}, we can see that incorporating a 12-hour forcing history already significantly improves peak performance of the baseline ST-GNN model compared to the simple GNN baseline that uses no historical forcing, highlighting the importance of historical forcing information for accurately predicting extreme surge levels. Our proposed PACT with the base configuration consistently outperforms the baseline on all peak-focused metrics, demonstrating the effectiveness of our model architecture and training approach in improving peak prediction. The improvements in Peak Mean Signed Error and Peak Max Absolute Error suggest that our peak-aware design not only reduces overall error but also helps mitigate systematic under-prediction of peaks and reduces the magnitude of the largest errors during extreme events. Notably, the best PACT configuration, which incorporates our peak-aware training objective, yields additional improvements in peak performance. This indicates that our peak-aware training objective is effective in further enhancing the model's ability to predict extreme surge levels beyond the improvements already achieved by the base configuration. Having examined both average performance and peak-event behavior, we next consider the practical computational efficiency of the proposed model.

\begin{table}[ht]
\centering
\caption{5\% Peak time performance comparison between our proposed PACT and baseline models on NCEP Reanalysis dataset at four different stations. Each station is trained and evaluated independently. Numbers are in meters\label{table:5-peak}}
\scalebox{0.7}{
\begin{tabular}{cccccc}
Stations                 & Metrics                 & \begin{tabular}[c]{@{}c@{}}Simple GNN Baseline\\ (0h Historical Forcing)\end{tabular} & \begin{tabular}[c]{@{}c@{}}ST-GNN Baseline\\ (12h Historical Forcing)\end{tabular} & \begin{tabular}[c]{@{}c@{}}PACT\\ (Base Config, Ours)\end{tabular} & \begin{tabular}[c]{@{}c@{}}PACT\\ (Best Config, Ours)\end{tabular} \\ \hline
\multirow{4}{*}{Battery} & Peak RMSE               & 0.1287                                                                     & 0.1022                                                                      & 0.0532                                                             & \textbf{0.0479}                                                    \\
                         & Peak MAE                & 0.0994                                                                     & 0.0796                                                                      & 0.0404                                                             & \textbf{0.0358}                                                    \\
                         & Peak Mean Signed Error  & -0.0893                                                                    & -0.0620                                                                     & -0.0195                                                            & \textbf{-0.0139}                                                   \\
                         & Peak Max Absolute Error & 0.4228                                                                     & 0.3537                                                                      & 0.2346                                                             & \textbf{0.2101}                                                    \\ \hline
\multirow{4}{*}{Boston}  & Peak RMSE               & 0.1021                                                                     & 0.0805                                                                      & 0.0441                                                             & \textbf{0.0413}                                                    \\
                         & Peak MAE                & 0.0765                                                                     & 0.0604                                                                      & 0.0332                                                             & \textbf{0.0313}                                                    \\
                         & Peak Mean Signed Error  & -0.0678                                                                    & -0.0395                                                                     & -0.0108                                                            & \textbf{-0.0075}                                                   \\
                         & Peak Max Absolute Error & 0.4446                                                                     & 0.4149                                                                      & 0.2050                                                             & \textbf{0.1881}                                                    \\ \hline
\multirow{4}{*}{CBBT}    & Peak RMSE               & 0.1078                                                                     & 0.0765                                                                      & 0.0397                                                             & \textbf{0.0375}                                                    \\
                         & Peak MAE                & 0.0911                                                                     & 0.0649                                                                      & 0.0312                                                             & \textbf{0.0293}                                                    \\
                         & Peak Mean Signed Error  & -0.0844                                                                    & -0.0542                                                                     & -0.0152                                                            & \textbf{-0.0134}                                                   \\
                         & Peak Max Absolute Error & 0.2959                                                                     & 0.2225                                                                      & 0.1446                                                             & \textbf{0.1320}                                                    \\ \hline
\multirow{4}{*}{Lewes}   & Peak RMSE               & 0.1225                                                                     & 0.0949                                                                      & 0.0464                                                             & \textbf{0.0434}                                                    \\
                         & Peak MAE                & 0.0977                                                                     & 0.0754                                                                      & 0.0352                                                             & \textbf{0.0330}                                                    \\
                         & Peak Mean Signed Error  & -0.0922                                                                    & -0.0629                                                                     & -0.0128                                                            & \textbf{-0.0091}                                                   \\
                         & Peak Max Absolute Error & 0.4281                                                                     & 0.3044                                                                      & \textbf{0.1792}                                                    & 0.1849                                                            
\end{tabular}
}
\end{table}

\subsection{Computational Efficiency}

Beyond prediction quality, an important motivation for learning-based surge emulation is speed; accordingly, we next assess the runtime characteristics of the proposed model, PACT with the best configurtaion. The training time is measured as the wall-clock time required to train the model on four NVIDIA H100 GPUs for 300 epochs. The inference time is measured as the average time taken to generate a complete surge-level trajectory for the winter season of a single year (from November 1st to March 30th), calculated by averaging the time taken for the model to make predictions across all available years in the dataset. 

\begin{table}[ht]
\centering
\caption{Runtime of our proposed PACT model with the best configuration and ADCIRC simulation on different datasets for one year.\label{table:training_inference_time}}
\begin{tabular}{cccc}
\hline
Dataset & Training Time(Seconds) & Inference Time(Seconds) & ADCIRC Simulation Time \\ 
\hline
NCEP     & 621  & 3.60 & \multirow{6}{*}{\parbox{2.2cm}{\centering 4.5--7 hours}} \\
AWI      & 2294 & 3.46 &  \\
CNRM     & 1214 & 3.40 &  \\
EC\_EARTH & 4452 & 3.54 &  \\
MPI      & 2292 & 3.50 &  \\
MRI      & 1531 & 3.45 &  \\
\hline
\end{tabular}
\end{table}

Table~\ref{table:training_inference_time} summarizes the training and inference times of PACT across the evaluated datasets. The training time varies across datasets, ranging from 621 seconds for NCEP to 4452 seconds for EC\_EARTH. By contrast, the inference time is relatively stable, averaging about 3.5 seconds to generate a full-year surge trajectory for a single dataset--station pair. In comparison, ADCIRC simulations typically require several hours to complete for a single year, depending on mesh resolution and available computational resources. This large gap highlights the computational advantage of PACT over traditional physics-based simulation. Although training requires a larger one-time cost, this cost can be amortized over repeated inference runs across multiple years, stations, or scenario ensembles. The short inference time is especially attractive for applications requiring rapid evaluation, such as large-scale scenario analysis or time-sensitive forecasting support. Taken together with the accuracy results above, the runtime comparison shows that PACT provides both strong predictive performance and substantial computational efficiency as a surrogate model for storm-surge emulation.
\section{Discussion}
\label{sec:discussion}

In Section~\ref{sec:results}, we showed that PACT consistently outperforms a baseline on the NCEP reanalysis dataset and that its peak-aware design further improves performance on metrics that emphasize extreme events. These findings suggest that the combination of station-conditioned spatial aggregation, horizon-aware temporal decoding, and peak-aware learning provides a more effective surrogate mapping from local atmospheric forcing to station-level surge response than a standard graph-based emulation architecture.

In this section, we discuss what these findings imply under more challenging and practically relevant settings, particularly when the model trained on one dataset needs to be applied to different forcing datasets for probabilistic climate scenario analysis. Using the CMIP6 datasets, we first examine in-dataset generalization, in which training, validation, and testing are all performed on the same individual CMIP6 forcing dataset. We then discuss cross-dataset transfer, where the model is trained on one forcing source and evaluated on another. We finally analyze peak-focused performance to assess how reliably PACT captures surge extremes. Together, these discussions help clarify both the strengths of PACT and the remaining challenges of storm-surge emulation under heterogeneous climate forcings.

\subsection{In-Dataset Generalization Ability}
We first consider the in-dataset setting, in which each model is trained, validated, and tested on the same individual CMIP6 forcing dataset rather than transferred across different forcing sources. To examine this setting under different temporal regimes, we consider two evaluation configurations: \textbf{Future-Period} and \textbf{Past-Only} (Section \ref{sec4.2}). 

\begin{table}[!ht]
\centering
\caption{In-dataset generalization performance on the CMIP6 dataset for past-only prediction. Only the past years of the CMIP6 dataset(1979-2014) were used. PACT is trained/validated/tested on the same CMIP6 dataset(AWI, CNRM, EC\_EARTH, MPI, or MRI). Metrics are reported in physical units (meters) as aggregated test-set RMSE and MAE, for the Battery, NY.\label{table:in-dataset-past}}
\scalebox{0.6}{
\begin{tabular}{ccccccccccc}
\multirow{2}{*}{Methods}                                                                   & \multicolumn{2}{c}{CMIP6-AWI} & \multicolumn{2}{c}{CMIP6-CNRM} & \multicolumn{2}{c}{CMIP6-EC\_EARTH} & \multicolumn{2}{c}{CMIP6-MPI} & \multicolumn{2}{c}{CMIP6-MRI} \\ \cline{2-11} 
                                                                                           & RMSE          & MAE           & RMSE           & MAE           & RMSE             & MAE              & RMSE          & MAE           & RMSE          & MAE           \\ \hline
\begin{tabular}[c]{@{}c@{}}ST-GNN Baseline\\ (12h Historical Forcing)\end{tabular}                & 0.0862        & 0.0612        & 0.0737         & 0.0533        &     0.0904         &  0.0649           & 0.0902        & 0.0645        & 0.0878        & 0.0628        \\
\begin{tabular}[c]{@{}c@{}}PACT\\ (12h Historical Forcing, Ours)\end{tabular} & 0.0422        & \textbf{0.0305}        & 0.0378         & 0.0281        &   0.0432           &    0.0315          & 0.0466        & 0.0334        & 0.0516        & 0.0350        \\
\begin{tabular}[c]{@{}c@{}}PACT + Peak Aware Loss\\ (12h Historical Forcing, Ours)\end{tabular} & \textbf{0.0421}        & 0.0306        & \textbf{0.0369}         & \textbf{0.0274}        &    \textbf{0.0417}          &      \textbf{0.0302}       & \textbf{0.0448}        & \textbf{0.0325}        & \textbf{0.0506}        & \textbf{0.0343}       
\end{tabular}
}
\end{table}

\begin{table}[!ht]
\centering
\caption{In-dataset generalization performance on CMIP6 dataset for future-period prediction. All the models are trained/validated/tested on the same CMIP6 dataset (AWI, CNRM, EC\_EARTH, MPI, or MRI), where the test set includes only future-period samples not seen during training/validation. Metrics are reported in physical units (meters) as aggregated test-set RMSE and MAE for the Battery, NY.\label{table:in-dataset-future}}
\scalebox{0.6}{
\begin{tabular}{ccccccccccc}
\multirow{2}{*}{Methods}                                                                   & \multicolumn{2}{c}{CMIP6-AWI} & \multicolumn{2}{c}{CMIP6-CNRM} & \multicolumn{2}{c}{CMIP6-EC\_EARTH} & \multicolumn{2}{c}{CMIP6-MPI} & \multicolumn{2}{c}{CMIP6-MRI} \\ \cline{2-11} 
                                                                                           & RMSE          & MAE           & RMSE           & MAE           & RMSE             & MAE              & RMSE          & MAE           & RMSE          & MAE           \\ \hline
\begin{tabular}[c]{@{}c@{}}ST-GNN Baseline\\ (12h Historical Forcing)\end{tabular}                & 0.0860        & 0.0609        & 0.0660         & 0.0480        &   0.0837           &   0.0601           & 0.0913        & 0.0650        & 0.0863        & 0.0619        \\
\begin{tabular}[c]{@{}c@{}}PACT\\ (Base Config, 12h Historical Forcing, Ours)\end{tabular} & 0.0406        & 0.0293        & 0.0357         & 0.0265        &   0.0452           &     0.0328       & 0.0446        & 0.0321        & 0.0437        & \textbf{0.0310}        \\
\begin{tabular}[c]{@{}c@{}}PACT + Peak Aware Loss\\ (Best Config, 12h Historical Forcing, Ours)\end{tabular} & \textbf{0.0402}        & \textbf{0.0288}        & \textbf{0.0348}         & \textbf{0.0258}        &   \textbf{0.0435}           &      \textbf{0.0311}       & \textbf{0.0442}        & \textbf{0.0318}        & \textbf{0.0437}        & 0.0311       
\end{tabular}
 }
\end{table}

Table~\ref{table:in-dataset-future} reports the \textbf{Future-Period} results. In this setting, we use only a single given CMIP6 dataset (AWI, CNRM, EC\_EARTH, MPI, or MRI), and PACT is trained on data from 1979-2008, validated on data from 2009-2014, and then tested on unseen future-period samples from 2070--2099. This setting therefore evaluates whether the learned forcing--surge relationship can generalize from historical climate conditions to a future climate regime within the same forcing source. Table~\ref{table:in-dataset-past} reports the \textbf{Past-Only} results. Here, training, validation, and testing are all restricted to the historical period of each CMIP6 dataset, using 1979--2000 for training, 2001--2007 for validation, and 2008--2014 for testing. This setting provides a within-dataset reference case without the shift from historical to future climate conditions.

Across all five CMIP6 GCM datasets and two evaluation settings, PACT consistently outperforms the baseline model under the same in-dataset training protocol. For future-period emulation, PACT reduces RMSE and MAE substantially for all the CMIP6 GCM datasets (roughly about 50\% relative reduction compared to the baseline). These gains suggest that the station-query readout and query-based temporal decoding provide a more robust mapping from spatial forcing patterns to station-level surge response, especially when the forcing statistics differ from typical historical conditions. For past-only prediction, PACT again yields large RMSE improvements. In addition, the best configuration of PACT consistently matches or slightly improves upon the base configuration across sources, with the largest differences appearing on the more difficult domains (e.g., EC\_EARTH), suggesting that the added peak-aware components primarily help under challenging tail behavior rather than uniformly shifting bulk accuracy. It is also interesting to note that for both versions of PACT, future-period emulation yields similar or better results when compared to the past-only generalization performance, despite the testing data being from a different climate regime than the training data in this case. The slightly longer training dataset length may be making up for the differences in the testing dataset. In addition, the errors shown for the Battery when training and testing both PACT versions on NCEP data, given in Table \ref{table:overall}, are lower than errors given by all GCM models for the Past-Only case. 

In summary, the in-dataset generalization results on CMIP6 datasets demonstrate that PACT's architectural designs provide substantial improvements over the baseline model across a range of data sources and time periods, with strong gains in both future-year and past-only prediction settings. The improvements are particularly pronounced for the more challenging EC\_EARTH dataset, indicating that PACT's design choices help the model better capture complex relationships between forcing and surge response under difficult conditions. These results highlight the potential of PACT as a powerful data-driven surrogate for storm-surge emulation across diverse climate-model outputs and temporal regimes. Taken together, these in-dataset results provide a reference point for interpreting the more demanding cross-dataset transfer behavior discussed next.

\subsection{Cross-Dataset Generalization Performance}

\begin{figure}[!ht]
  \centering
  \includegraphics[width=1.0\textwidth]{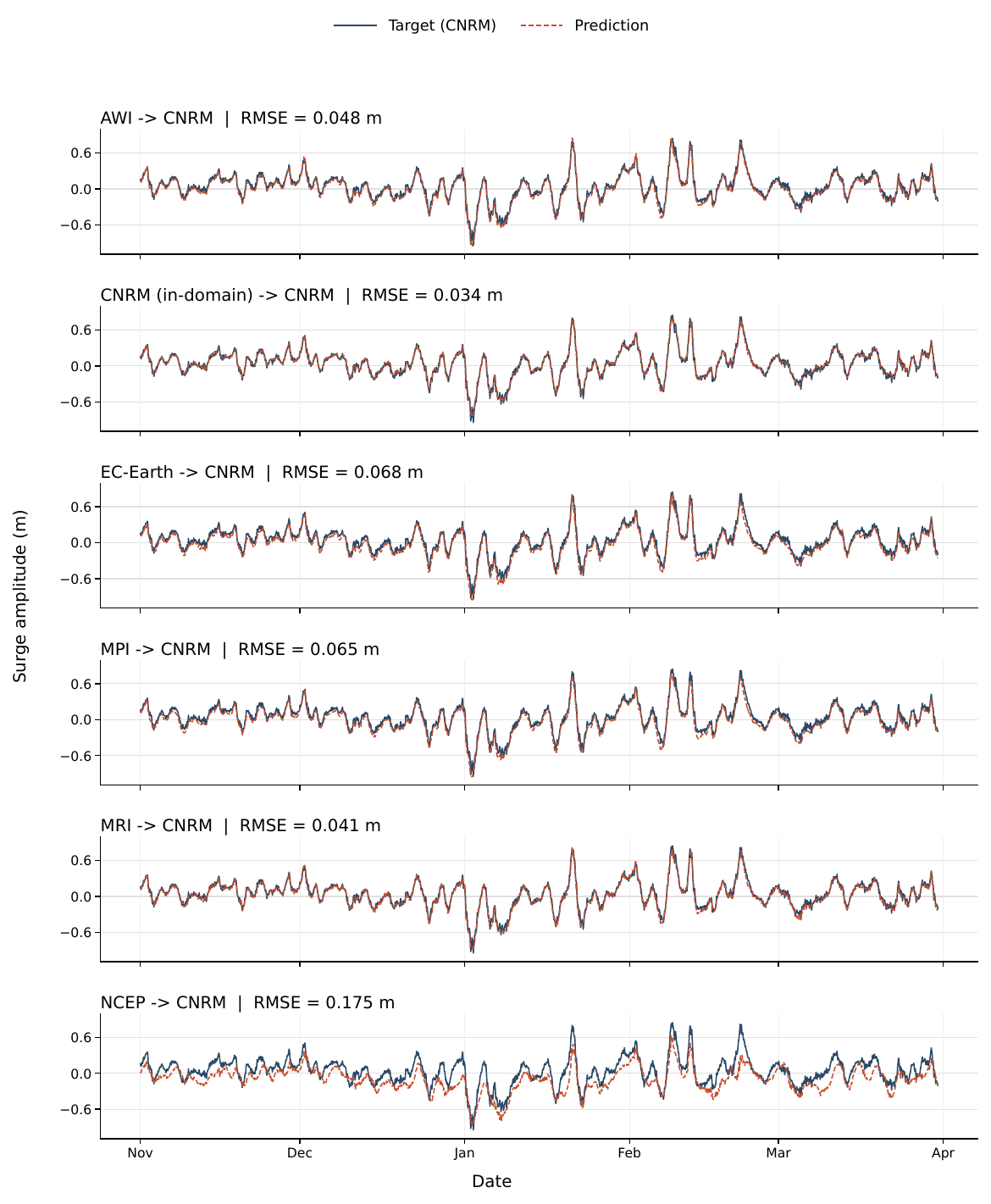}
  \caption{Qualitative examples of surge level time series predictions on year 2080-2081 in the CNRM dataset, with the model trained on different datasets.\label{fig:qual-cross-dataset}}
\end{figure}

Having established the within-dataset performance of PACT, we now examine cross-dataset generalization, which is substantially more demanding because the atmospheric forcing distributions can differ across reanalysis and climate-model products. 
We focus on the best configuration of PACT (with a peak-aware loss) and evaluate cross-dataset transfer by training PACT on one forcing dataset, referred to as the source dataset, and applying the trained model directly to a different forcing dataset, referred to as the target dataset, without retraining. Because NCEP is available only for the historical period, whereas the five CMIP6 GCM datasets include both historical and future periods, we report two complementary cross-dataset settings.

In the \textbf{Past-Only} setting, all source models are trained using the historical split of the source dataset, with 1979--2000 for training and 2001--2007 for validation. The trained model is then evaluated on the historical period, 1979--2014, of each different target dataset.

In the \textbf{Future-Period} setting, the target evaluation is restricted to the CMIP6 future period, 2070--2099. When the source dataset is NCEP, the model is trained on 1979--2000 and validated on 2001--2007. When the source dataset is CMIP6, the model is trained on 1979--2008 and validated on 2009--2014. In all cases, evaluation is performed on the future period of a different CMIP6 target dataset.

Tables~\ref{table:cross-dataset-past-rmse} and  \ref{table:cross-dataset-future-rmse} report the RMSE results for the Past-Only and Future-Period, respectively. The corresponding MAE results are provided in the~\ref{appendix:additional-cross-dataset}. Across all results, models trained on NCEP exhibit substantially larger errors when transferred to any CMIP6 GCM dataset (Table \ref{table:cmip6}). This indicates a pronounced domain shift between reanalysis and GCM datasets, where differences in large-scale bias, variance, and the spatiotemporal structure of atmospheric fields can lead to mismatched surge responses if not explicitly accounted for. In contrast, cross-dataset transfer among the CMIP6 datasets is markedly better, comparable to the in-dataset performance shown in Table \ref{table:in-dataset-past}, exposing a meaningful similarity structure. Most GCM$\rightarrow$GCM transfers remain in the 0.04--0.09 RMSE range, suggesting that once trained within the GCM domain, PACT can exploit shared physical relationships across multiple climate models. At the same time, transfer is not uniformly symmetric: certain source--target pairs are consistently harder, with EC\_EARTH often acting as a challenging target, indicating forcing characteristics that are less well covered by models trained on other GCMs. EC\_EARTH represents the highest-resolution forcing dataset used in this study, and thus may introduce smaller-scale details that can be harder for the model to capture when trained on lower-resolution datasets. When the model is trained on coarser-resolution forcings, it primarily learns a mapping that reflects smoothed wind and pressure fields, and may not fully capture the influence of finer-scale gradients or localized forcing structures. As a result, when evaluated on EC\_EARTH, the model is exposed to higher-frequency variability and sharper forcing features that fall outside the training distribution.

A possible explanation for the reanalysis$\rightarrow$GCM degradation is a domain-shift effect rather than a failure of the emulator architecture. NCEP reanalysis is constrained by data assimilation that reflects the observed atmospheric state and tends to exhibit distinct bias/variance characteristics and spatiotemporal spectra compared to GCM datasets, which are not observation-constrained and can differ systematically in storm statistics, background climatology, and extremes. Importantly, all reported errors are measured against ADCIRC-generated surge targets obtained by driving the simulator with the corresponding forcing source. Therefore, the large NCEP$\rightarrow$GCM degradation should not be interpreted as evidence that GCM forcings are ``less accurate'' in an observational sense. Instead, it reflects a mismatch between source-induced input--output relationships: a model trained to emulate the mapping defined by NCEP-driven simulations does not, in general, transfer to the distinct mapping induced by a particular GCM forcing distribution. This systematic difference in how reanalysis and GCM data resolve ETCs has also been noted in prior work, which found that CMIP6 models struggle to reproduce specific patterns in ETC dynamics. Specifically, CMIP6 models tend toward weaker cyclones that travel zonally and underestimate both the frequency and rapid intensification rates of high-intensity "bomb" cyclones relative to reanalysis products~\cite{priestley2020,gore2023}. At the same time, the comparatively strong GCM$\leftrightarrow$GCM transfer indicates that the dominant obstacle is the reanalysis--GCM distribution gap, rather than an inherent limitation of the architecture's ability to generalize within the CMIP6 family. These findings motivate two practical directions for improving robustness under forcing shift: multi-source training across multiple GCMs to broaden coverage of forcing variability, and lightweight domain adaptation or calibration when deploying models across reanalysis and climate-model forcings, where distributional differences can substantially alter the effective simulator response.

We also observe an asymmetry between reanalysis and GCM domains: while NCEP$\rightarrow$GCM transfer is generally poor, GCM-trained models can transfer back to NCEP with comparatively smaller errors, consistent with the view that the main difficulty is the reanalysis--climate distribution gap rather than the inability to generalize within the broader CMIP6 family. This asymmetry is consistent with a ``harder-to-easier'' transfer direction: CMIP6 forcings are free-running and typically less observation-constrained, and their forcing fields may exhibit larger biases, different variance, and altered storm statistics relative to reanalysis. Training on such a distribution can implicitly expose the model to a broader and noisier set of input patterns, which may encourage learning a more conservative, regularized input--output mapping that still performs well when evaluated on the more observation-consistent NCEP regime. In this sense, transferring from the GCM domain to NCEP resembles moving from a higher-variability (and potentially less reliable) forcing space to a more stable and accurate reference distribution. Conversely, a model fit tightly to the NCEP-induced mapping may rely on statistical relationships that do not persist under GCM forcings, leading to large degradation when transferring in the opposite direction. For the future-period cross-data evaluation (Table \ref{table:cross-dataset-future-rmse}), the RMSE values are only moderately increased compared to the RMSEs for in-data generalization (Table \ref{table:in-dataset-future}), and exhibit a similar pattern of strong GCM$\leftrightarrow$GCM transfer but weak NCEP$\rightarrow$GCM transfer. In addition, this case results in similar or even better performance when compared to the past-only evaluation (Table \ref{table:cross-dataset-past-rmse} vs. Table \ref{table:cross-dataset-future-rmse}), most likely due to the longer training record. This indicates that PACT remains robust when extrapolating to future climate conditions. 

Overall, the results show strong transfer among GCM datasets and from GCMs to NCEP, but substantially weaker transfer from NCEP to GCMs. This asymmetric behavior suggests that models trained on reanalysis forcing do not fully capture the variability present in climate-model forcing, making the reanalysis-to-GCM-domain shift the primary barrier to out-of-domain robustness. To illustrate, Figure~\ref{fig:qual-cross-dataset} shows examples of surge-level time-series predictions for 2080--2081 in the CNRM dataset, using models trained on different datasets. Across GCM$\rightarrow$CNRM transfers, the predicted hourly surge closely tracks the ADCIRC target in both phase and amplitude, including short-lived spikes and multi-day oscillations, indicating that the learned input--output relationship generalizes reasonably within the CMIP6 family. In contrast, the NCEP$\rightarrow$CNRM transfer exhibits visibly attenuated peak amplitudes and a more persistent negative deviation during energetic periods, consistent with a distribution-induced mapping mismatch between reanalysis-driven and GCM-driven simulations. These findings suggest that while PACT's architecture can generalize across a range of forcing distributions, the choice of training data is critical for ensuring robust performance under distributional shift. In particular, training on a diverse set of GCM forcings may help mitigate the domain gap and improve transferability to new datasets, while lightweight domain adaptation techniques may be necessary when transferring from reanalysis to climate-model forcings, given their distinct statistical properties. These insights are important for real-world applications of surge emulators, where the model may be deployed on new datasets with different forcing characteristics, and underscore the need to consider domain shift and transferability in the design and evaluation of learning-based surrogates for storm surge emulation.

\begin{table}[ht]
\centering
\caption{Past-only RMSE error cross-dataset evaluation results of PACT with the best configuration, trained on one dataset and evaluated on other datasets. Numbers are in meters, for the Battery, NY.\label{table:cross-dataset-past-rmse}}
\setlength{\tabcolsep}{6pt}
\renewcommand{\arraystretch}{1.15}
\begin{tabular}{c c c c c c c}
& \multicolumn{5}{c}{Cross-Model Evaluation Dataset} \\
\cmidrule{2-7}
\textbf{Train/Val} & NCEP & AWI & CNRM & EC\_EARTH & MPI & MRI \\
\midrule
NCEP      & / & 0.1616 & 0.1789 & 0.1425 & 0.1595 & 0.1779 \\
AWI       & 0.0534 & / & 0.0543 & 0.0548 & 0.0425 & 0.0519 \\
CNRM      & 0.0510 & 0.0868 & / & 0.1147 & 0.0886 & 0.0609 \\
EC\_EARTH & 0.0575 & 0.0472 & 0.0722 & / & 0.0492 & 0.0632 \\
MPI       & 0.0814 & 0.0419 & 0.0689 & 0.0591 & / & 0.0565 \\
MRI       & 0.0630 & 0.0481 & 0.0440 & 0.0761 & 0.0509  & / \\
\end{tabular}
\end{table}

\begin{table}[ht]
\centering
\caption{Future-period RMSE error cross-dataset evaluation results of our proposed PACT with best configuration, trained on one dataset and evaluated on other datasets. "n.a." is used for evaluations on NCEP as it only includes past years. Numbers are in meters.\label{table:cross-dataset-future-rmse}}
\setlength{\tabcolsep}{6pt}
\renewcommand{\arraystretch}{1.15}
\begin{tabular}{c c c c c c c}
& \multicolumn{5}{c}{Cross-Model Evaluation Dataset} \\
\cmidrule{2-7}
\textbf{Train/Val} & NCEP & AWI & CNRM & EC\_EARTH & MPI & MRI \\
\midrule
NCEP      & / & 0.1506 & 0.1724 & 0.1378 & 0.1574 & 0.1768 \\
AWI       & n.a. & / & 0.0506 & 0.0567 & 0.0437 & 0.0506 \\
CNRM      & n.a. & 0.0817 & / & 0.1115 & 0.0858 & 0.0563 \\
EC\_EARTH & n.a. & 0.0472 & 0.0693 & / & 0.0501 & 0.0649 \\
MPI       & n.a. & 0.0416 & 0.0654 & 0.0596 & / & 0.0561 \\
MRI       & n.a. & 0.0471 & 0.0421 & 0.0761 & 0.0522  & / \\
\end{tabular}
\end{table}

\subsection{Performance on Predicting Storm Surge Peaks}
\label{sec:peak_eval}

Because the most societally consequential surge events are concentrated in the upper tail of the distribution, it is important to examine not only overall error but also how well the model captures peak behavior. To evaluate PACT's performance in predicting storm-surge peaks, we conduct an event-based peak-intensity analysis using the model’s multi-horizon outputs alongside the corresponding ground-truth surge trajectories (ADCIRC outputs). For each station and year, the model produces a sequence of short-horizon predictions at hourly resolution over a fixed 6-hour lead-time window. We aggregate these predictions into a continuous hourly series that is temporally aligned with the hourly ground-truth series for the same period. This yields paired time series $\{y(t)\}_{t=1}^T$ and $\{\hat y(t)\}_{t=1}^T$ in physical units (meters). Here, we focus only on the performance of future-year emulation at the Battery station using the CMIP6 datasets. For each CMIP6 dataset (AWI, CNRM, EC\_EARTH, MPI, or MRI), we train, validate, and test the model within the same dataset.

Storm-surge peaks are then extracted at the event level by exceeding a high-quantile threshold. For each reconstructed yearly series, we compute a threshold
\begin{equation}
u_q=\mathrm{Quantile}_q(\{y(t)\}_{t=1}^T),
\end{equation}
with $q=0.95$ in our experiments, and identify exceedance hours $\mathcal{I}=\{t:\,y(t)>u_q\}$. We cluster exceedances into events by splitting $\mathcal{I}$ whenever the gap between consecutive exceedances exceeds a minimum separation of 24 hours in our setting. To suppress spurious detections, we discard clusters shorter than a minimum 3-hour exceedance duration. For each retained event cluster $\mathcal{C}$, we record the event window and define the event peak as
\begin{equation}
p=\max_{t\in\mathcal{C}} y(t), \qquad t^\star=\arg\max_{t\in\mathcal{C}} y(t).
\end{equation}
The same procedure is applied independently to the predicted series $\hat y(t)$, producing predicted peak values and peak times. 

Because peaks are extracted independently from $y(t)$ and $\hat y(t)$, the number of detected events can differ between ground truth and predictions. To obtain a simple and reproducible pairing, we pair events by their time-order index within each year: the $i$-th predicted event is paired with the $i$-th ground-truth event, up to $m=\min(n_{\mathrm{gt}},n_{\mathrm{pred}})$. We further filter pairs by requiring that the absolute difference in peak times be within 48 hours when timestamps are available, thereby removing gross temporal mismatches. Paired peaks are then pooled across all the future years to compute peak-centric diagnostics.

Using the paired peak set $\{(p_i,\hat p_i)\}$, we report three complementary evaluations. First, we visualize peaks using a scatter plot of predicted versus ground-truth peaks against the identity line, where systematic deviations indicate under- or over-prediction at high water levels. Second, we assess distributional fidelity by comparing empirical peak distributions using a Gaussian kernel density estimate with a shared bandwidth to ensure comparability across methods; this evaluates whether a model preserves tail heaviness and allocates sufficient probability mass to large peaks. Third, we quantify how error scales with event severity by binning ground-truth peak magnitudes into quantile bins spanning the 1st to the 99th percentiles and computing RMSE within each bin. To mitigate sampling noise, we smooth the resulting per-bin RMSE curve using a short moving-window mean and exclude bins with fewer than a specified minimum number of events. Together, these analyses characterize not only average peak accuracy, but also tail distribution and the growth of error under increasingly extreme events.

\begin{figure}[!ht]
  \centering
  \includegraphics[width=1.0\textwidth]{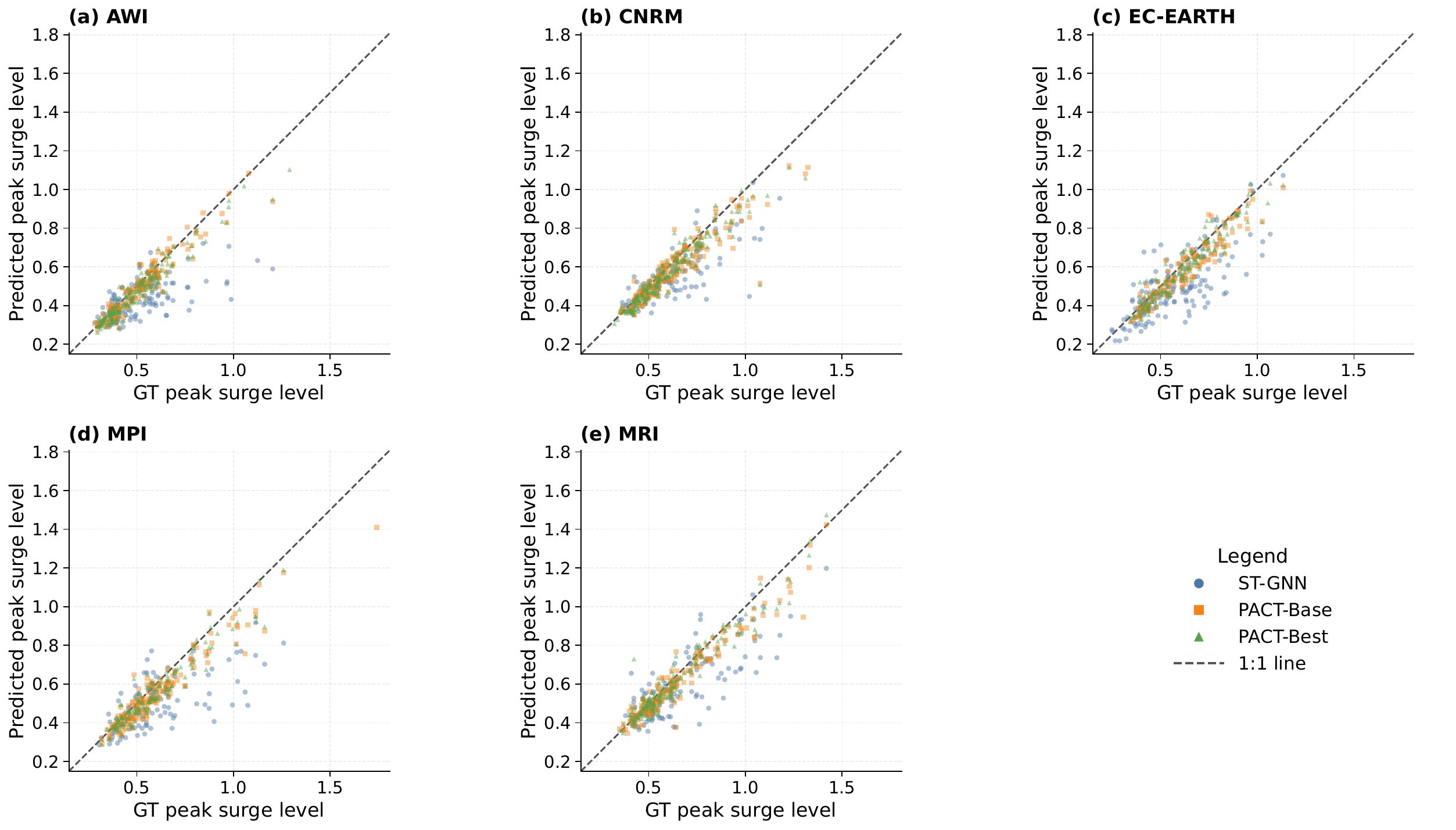}
  \caption{Event-level peak prediction accuracy on the Battery station for future-year CMIP6 forcings. Each point corresponds to a detected surge event and plots the predicted peak magnitude $\hat p$ against the ground-truth peak magnitude $p$ (in meters). Events are identified as clusters of exceedances above the 95th percentile within each year, and predicted/ground-truth events are paired in temporal order with a 48-hour peak-time tolerance. Colors compare Baseline-12h, PACT-Base, and PACT-Best; the dashed line indicates perfect agreement ($\hat p=p$). Panels correspond to different CMIP6 datasets (AWI, CNRM, EC\_EARTH, MPI, MRI).\label{fig:peak_scatter}}
\end{figure}

Figure~\ref{fig:peak_scatter} plots predicted peak surge against the ground-truth peak for each CMIP6 forcing dataset. The baseline exhibits broader dispersion and a systematic tendency to fall below the $y=x$ line at larger peaks, indicating underprediction in high-impact regimes. The PACT variants, especially the PACT-Best, tighten the scatter around the diagonal, reducing heteroscedasticity as peak magnitude increases and improving peak-prediction accuracy in the upper tail. Overall, this figure supports the claim that PACT improves event-level peak fidelity—not merely average error—by reducing systematic peak underestimation and improving peak-wise consistency across forcings.

\begin{figure}[!ht]
  \centering
  \includegraphics[width=1.0\textwidth]{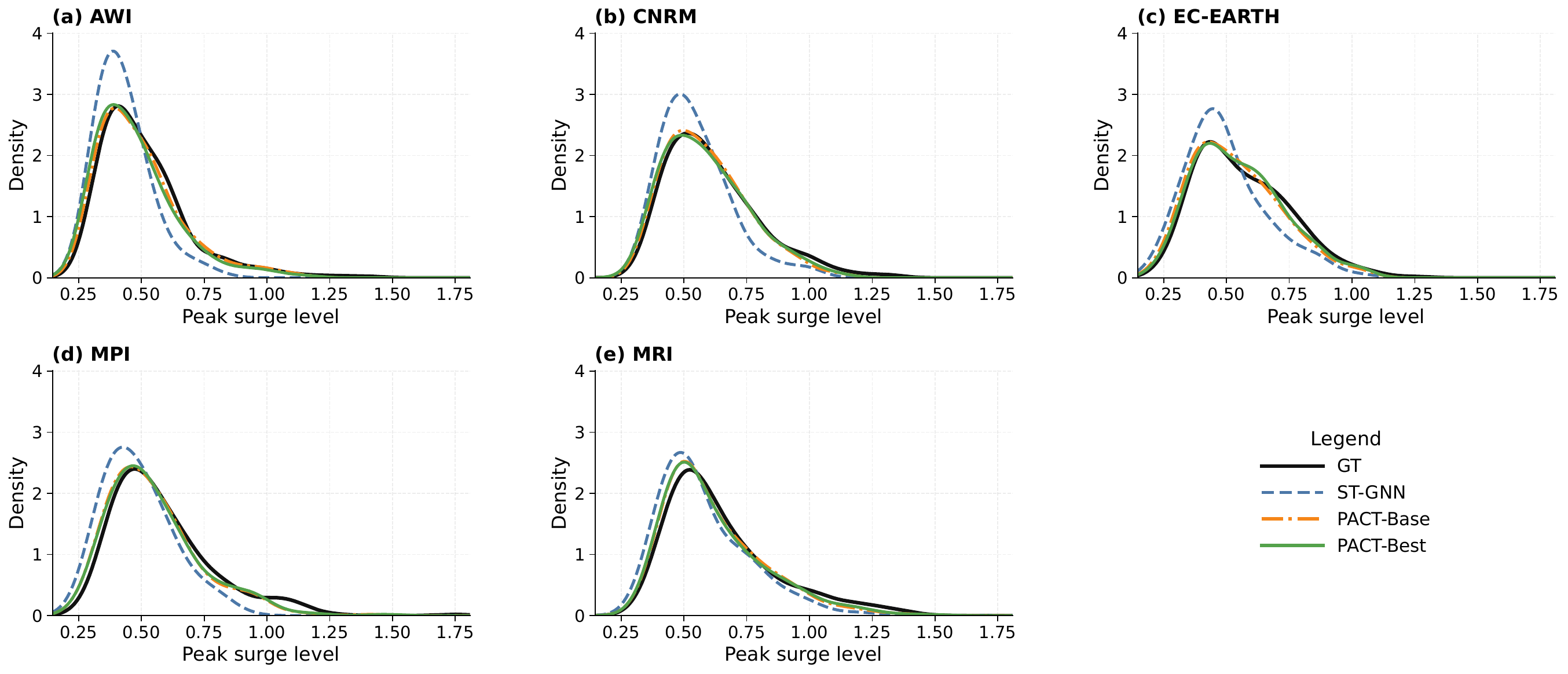}
  \caption{Distributional fidelity of storm-surge peak magnitudes on the Battery station for future-year CMIP6 forcings. We report Gaussian kernel density estimates of event peak magnitudes (meters) for the ADCIRC-driven ground truth (GT) and model predictions (Baseline-12h, PACT-Base, PACT-Best), using a shared bandwidth for comparability. Panels correspond to different CMIP6 datasets (AWI, CNRM, EC\_EARTH, MPI, MRI).\label{fig:peak_density}}
\end{figure}

Figure~\ref{fig:peak_density} evaluates whether models reproduce the distribution of peak surge values, which is crucial for downstream risk metrics (e.g., exceedance probabilities and return levels). Across datasets, the baseline density is typically over-concentrated around moderate peaks and decays too quickly in the upper tail, consistent with peak damping and under-dispersion. PACT produces density curves that more closely follow the ADCIRC-driven ground-truth distribution, especially in the shoulder-to-tail transition, indicating improved allocation of probability mass to larger peaks. Among the evaluated forcings, EC\_EARTH is the most difficult to reproduce in the upper tail: PACT improves the shoulder-to-tail transition relative to the baseline, but the density of the largest peaks remains underestimated.

\begin{figure}[!ht]
  \centering
  \includegraphics[width=1.0\textwidth]{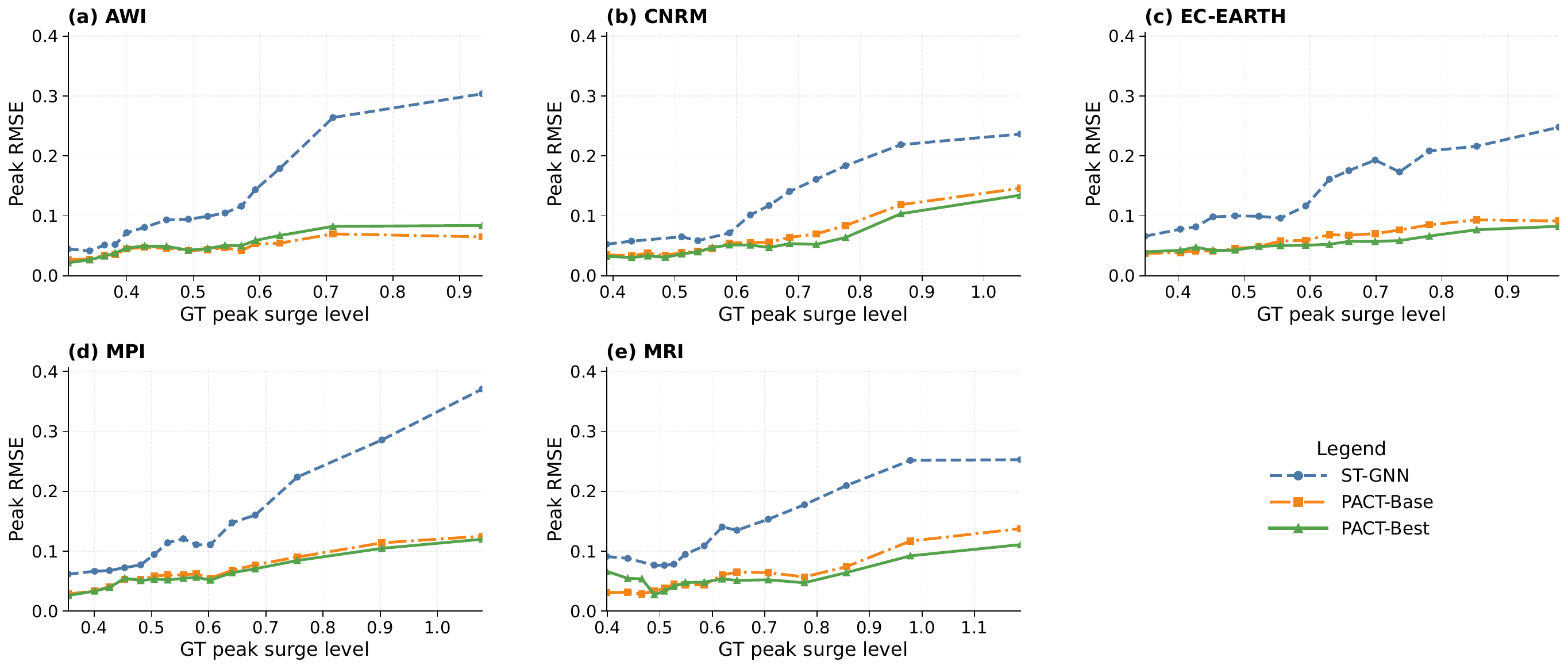}
  \caption{Severity-conditioned peak error on the Battery station for future-year CMIP6 forcings. Peak RMSE (meters) is computed as a function of event severity by binning ground-truth peak magnitudes into quantile bins spanning the 1st--99th percentiles and evaluating RMSE within each bin. The resulting curve is smoothed with a short moving-window mean, and bins with fewer than five events are omitted. Panels correspond to different CMIP6 datasets (AWI, CNRM, EC\_EARTH, MPI, MRI).\label{fig:peak_rmse_vs_level}}
\end{figure}

Figure~\ref{fig:peak_rmse_vs_level} reports peak RMSE conditioned on the ground-truth peak magnitude. The baseline error increases sharply as events become more severe, indicating that its mistakes are highly event-dependent and disproportionately concentrated in the tail. In contrast, both PACT variants substantially flatten this error-growth curve: peak RMSE remains comparatively stable through moderate-to-large peaks and rises more gradually at the extreme end. This pattern is consistent with a model that learns a more robust mapping for rare, high-amplitude events rather than optimizing primarily for the bulk of the distribution. Figure~\ref{fig:peak_rmse_vs_level} shows a clear improvement for all the datasets that PACT-Best, which incorporates the peak-aware loss, further suppresses error growth at the upper end of the peak spectrum compared to PACT-Base, indicating that explicitly emphasizing peak samples during training can enhance tail performance. Overall, the key takeaway is that PACT reduces tail-amplified errors, a critical source of degradation for operational surge prediction.

\section{Conclusion}

In this work, we presented PACT, a peak-aware storm-surge emulator designed to improve prediction fidelity in both typical conditions and high-impact extremes for forcing-driven, season-scale, multi-trajectory emulation problems. PACT is a spatio-temporal graph Transformer that combines spatial cross-attention, temporal self-attention, and horizon-query decoding for multi-step surge prediction. Across multiple tide-gauge stations on NCEP reanalysis, PACT substantially reduced RMSE and MAE relative to a baseline spatio-temporal graph neural network model, and simple physics-motivated preprocessing (spatial mean centering of pressure) provided additional consistent gains. To better target rare but consequential events, we introduced a peak-aware training objective that augments standard MSE with a tail-focused loss over peak-dominated samples and a horizon-wise slope regularizer that encourages physically plausible multi-horizon trajectories. Ablations show that these components are complementary: the tail term strengthens peak-intensity learning and reduces worst-case peak errors, while the slope term improves trajectory consistency and yields a more balanced trade-off between bulk and tail metrics. In terms of computational efficiency, PACT is lightweight at inference: generating a full winter-season surge trajectory for a single year (Nov~1--Mar~30) takes roughly 3.4--3.6 seconds on average across datasets. Training on four NVIDIA H100 GPUs for 300 epochs completes in 621--4452 seconds, depending on the forcing source, making the one-time training cost practical relative to the large number of trajectories produced during deployment.

We further evaluate PACT's generalization capabilities under distribution shift, considering both in-dataset generalization and cross-dataset transfer. For in-dataset generalization, PACT maintains strong performance when trained and evaluated on the CMIP6 GCM datasets instead of the observation-constrained NCEP reanalysis dataset, indicating that our proposed method is robust to different data sources and does not rely on narrow, year-specific correlations. We then study cross-dataset robustness by transferring models between the NCEP reanalysis and five CMIP6 GCM datasets (AWI, CNRM, EC\_EARTH, MPI, and MRI). The results show consistently strong transfer within the CMIP6 family and CMIP6$\leftrightarrow$NCEP transfer, whereas transferring from NCEP reanalysis to CMIP6 GCM datasets leads to a pronounced error increase, highlighting the reanalysis--GCM distribution gap as a primary barrier to out-of-domain generalization. 

Finally, we assess peak fidelity on different CMIP6 future-year emulation using peak-centric diagnostics. Results show that introducing the peak-aware objective improves peak-intensity prediction and better preserves tail behavior across CMIP6 datasets. Nevertheless, this study demonstrates that PACT can preserve temporal evolution and risk-relevant tail behavior of storm surges, and introduces an architecture that can potentially be applied to further geophysical problems that deal with localized responses to gridded environmental forcing. 

These findings suggest several directions for future work and model improvement. First, multi-source training across diverse GCM forcings and scenarios may improve coverage of forcing variability and reduce hard-transfer failures. Second, lightweight adaptation or calibration schemes, particularly those that explicitly control tail intensity, appear promising for stabilizing return-level extrapolation under strong forcing shift. Third, the incorporation of station metadata that drive surge effects, such as depths, bathymetric gradients, and coastline orientation angle, as fixed station features can be explored as a means to improve performance. Fourth, we can test the performance of alternative loss functions, such as the Weighted Quantile and Expectile loss (WQE) introduced by~\cite{longo2026}, which was shown to reduce underestimation of extreme values in storm surge emulation models. Finally, extending the peak-aware objective to incorporate uncertainty quantification and probabilistic EVT-consistent losses could further improve reliability for operational flood-risk assessment.

\appendix
\section{Architecture Diagram of the ST-GNN Baseline Model}
\label{appendix: ST-GNN}

\begin{figure}[h]
  \centering
  \includegraphics[width=0.5\textwidth]{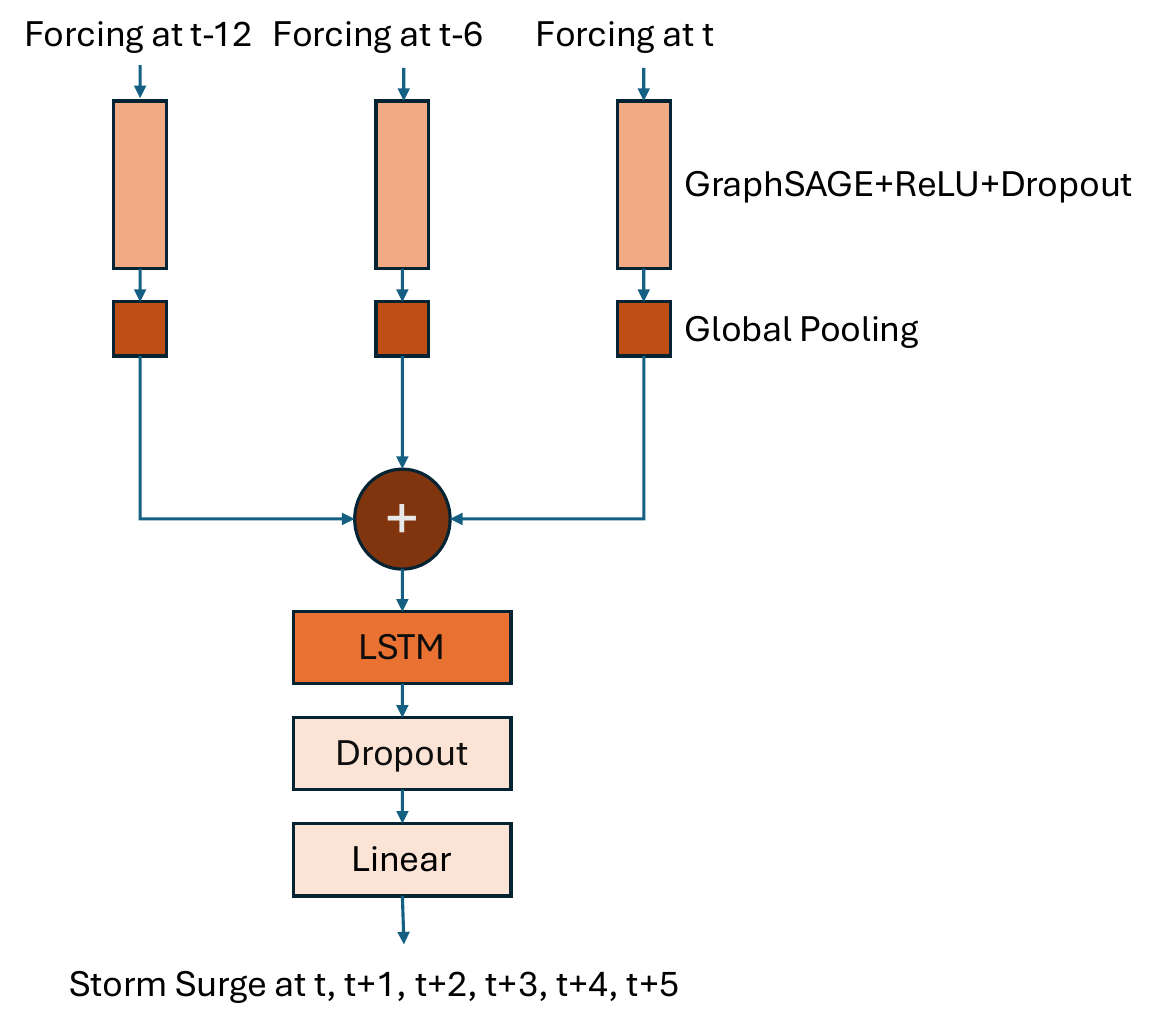}
  \caption{Architecture Diagram of the ST-GNN baseline model used in this work.\label{fig:STGNN}}
\end{figure}

Figure~\ref{fig:STGNN} shows the architecture of the ST-GNN baseline model used in this work, which is composed of GraphSAGE, global pooling and long short-term memory(LSTM).

\section{Ablation Study on Spatial Mean Centering for Pressure Anomalies}
\label{appendix: ablation-spatial-mean}

Table~\ref{table:spatial-mean-removal} evaluates the effect of spatial mean centering applied to the pressure forcing field. Across all four stations, removing the spatial mean yields modest improvements in overall error. We found that for certain stations, such as Battery, the improvement is more pronounced, with RMSE and MAE reductions of 0.0006m and 0.0005m, respectively. For other stations, the improvements are smaller but still consistent, with RMSE reductions of 0.0001--0.0002m and MAE reductions of up to 0.0005m. This trend supports the intuition that storm-surge response is driven more by spatial pressure gradients and synoptic-scale anomalies than by the absolute pressure offset, and that centering the pressure field provides a mild but reliable inductive bias for learning these anomaly-driven relationships.

\begin{table}[ht]
\centering
\caption{Ablation of spatial mean centering for the pressure feature in PACT (base configuration) on the NCEP Reanalysis dataset. Each station is trained and evaluated independently. We report RMSE and MAE in meters (lower is better). Spatial mean centering is applied per time step over the forcing grid to emphasize pressure anomalies rather than absolute pressure levels.\label{table:spatial-mean-removal}}
\scalebox{0.7}{
\begin{tabular}{ccccccccc}
\multirow{2}{*}{Methods}                  & \multicolumn{2}{c}{Battery} & \multicolumn{2}{c}{Boston} & \multicolumn{2}{c}{Lewes} & \multicolumn{2}{c}{CBBT} \\ \cline{2-9} 
                                          & RMSE         & MAE          & RMSE         & MAE         & RMSE        & MAE         & RMSE        & MAE        \\ \hline
With Spatial Mean Removal for Pressure    & \textbf{0.0349}       & \textbf{0.0253}       & \textbf{0.0282}       & \textbf{0.0209}      & \textbf{0.0279}      & \textbf{0.0210}      & \textbf{0.0309}      & \textbf{0.0237}     \\
Without Spatial Mean Removal for Pressure & 0.0355       & 0.0258       & 0.0284       & 0.0210      & 0.0280      & 0.0210      & 0.0310      & 0.0238    
\end{tabular}
}
\end{table}

\section{Ablation Study on Peak-Aware Loss Design}
\label{appendix:ablation-loss}

\begin{table}[ht]
\centering
\caption{Ablation of peak-aware loss components for PACT on the Battery station and NCEP Reanalysis dataset. We compare training with standard MSE against adding a tail-focused term defined on the top-$\rho$ peak samples (via a fixed training-set threshold $\tau_{\text{tail}}$) and a horizon-wise slope-matching regularizer that aligns first differences across lead times using a robust Charbonnier penalty. We report overall RMSE/MAE and peak-centric metrics computed on subsets defined by the top 10\%, 5\%, and 1\% of ground-truth surge values (lower is better). Mean signed error (MSErr) indicates bias on peaks (negative implies underprediction). All metrics are in meters.\label{table:loss}}
\scalebox{0.7}{
\begin{tabular}{ccccc}
Metrics                      & MSE     & MSE + Tail       & MSE + Slope     & MSE + Tail + Slope \\ \hline
Overall RMSE                 & 0.0349  & 0.0345           & 0.0340          & \textbf{0.0337}    \\
Overall MAE                  & 0.0253  & 0.0252           & 0.0247          & \textbf{0.0246}    \\ \hline
10\% Peak RMSE               & 0.0450  & 0.0411           & 0.0430          & \textbf{0.0408}    \\
10\% Peak MAE                & 0.0334  & 0.0308           & 0.0320          & \textbf{0.0302}    \\
10\% Peak Mean Signed Error  & -0.0121 & -0.0078          & -0.0108         & \textbf{-0.0076}   \\
10\% Peak Max Absolute Error & 0.2346  & \textbf{0.1834}  & 0.2582          & 0.2101             \\ \hline
5\% Peak RMSE                & 0.0532  & \textbf{0.0477}  & 0.0510          & 0.0479             \\
5\% Peak MAE                 & 0.0404  & 0.0364           & 0.0387          & \textbf{0.0358}    \\
5\% Peak Mean Signed Error   & -0.0195 & \textbf{-0.0137} & -0.0176         & -0.0139            \\
5\% Peak Max Absolute Error  & 0.2346  & \textbf{0.1834}  & 0.2582          & 0.2101             \\ \hline
1\% Peak RMSE                & 0.0627  & 0.0559           & 0.0590          & \textbf{0.0556}    \\
1\% Peak MAE                 & 0.0505  & 0.0448           & 0.0478          & \textbf{0.0438}    \\
1\% Peak Mean Signed Error   & -0.0260 & \textbf{-0.0189} & -0.0234         & -0.0193            \\
1\% Peak Max Absolute Error  & 0.1718  & 0.1617           & \textbf{0.1500} & 0.1637            
\end{tabular}
}
\end{table}

Table~\ref{table:loss} isolates the contributions of the proposed peak-aware loss components. Relative to plain MSE training, adding either the tail loss or slope loss term improves peak performance, with the combined objective providing the most balanced gains across overall and peak-specific metrics. The tail-focused term yields the largest improvements on several peak-specific criteria, while the slope-matching term helps control worst-case behavior and encourages more physically plausible multi-horizon trajectories.

The two components exhibit complementary behavior consistent with their definitions.The tail term, which applies an additional MSE over samples satisfying $p_b\ge\tau_{\text{tail}}$, most directly amplifies the gradient signal on extreme storm events and therefore yields the greatest improvements on several tail-specific criteria, including the best 10\%/5\% peak max absolute error. In contrast, the slope term constrains horizon-to-horizon dynamics by matching first differences $\Delta \hat y_{b,h}$ to $\Delta y_{b,h}$, which improves global fidelity and helps control the most extreme worst-case behavior (best 1\% peak max absolute error). 

Although tail-only is optimal for certain individual peak metrics, we adopt the combined MSE + Tail + Slope objective as the default because it offers a superior Pareto trade-off: it delivers the best overall RMSE/MAE and consistently strong peak RMSE/MAE across multiple peak thresholds, while simultaneously reducing peak bias and discouraging implausibly jagged multi-horizon trajectories. This robustness is important in deployment settings where both bulk accuracy and reliable, physically plausible evolution during extremes are required, rather than optimizing a single tail statistic in isolation.

\section{Additional Peak-Evaluation Results at Top 1\% and Top 10\% Thresholds}
\label{appendix:additional-peak-thresholds}

\begin{table}[!t]
\centering
\caption{10\% Peak time performance comparison between our proposed PACT and baseline model on NCEP Reanalysis dataset at four different stations. Each station is trained and evaluated independently. Numbers are in meters.\label{table:10-peak}}
\scalebox{0.6}{
\begin{tabular}{cccccc}
Stations                 & Metrics                 & \begin{tabular}[c]{@{}c@{}}ST-GNN Baseline\\ (0h Historical Forcing)\end{tabular} & \begin{tabular}[c]{@{}c@{}}ST-GNN Baseline\\ (12h Historical Forcing)\end{tabular} & \begin{tabular}[c]{@{}c@{}}PACT\\ (Base Config, Ours)\end{tabular} & \begin{tabular}[c]{@{}c@{}}PACT\\ (Best Config, Ours)\end{tabular} \\ \hline
\multirow{4}{*}{Battery} & Peak RMSE               & 0.1010                                                                     & 0.0821                                                                      & 0.0450                                                             & \textbf{0.0408}                                                    \\
                         & Peak MAE                & 0.0740                                                                     & 0.0610                                                                      & 0.0334                                                             & \textbf{0.0302}                                                    \\
                         & Peak Mean Signed Error  & -0.0527                                                                    & -0.0402                                                                     & -0.0121                                                            & \textbf{-0.0076}                                                   \\
                         & Peak Max Absolute Error & 0.4228                                                                     & 0.3537                                                                      & 0.2346                                                             & \textbf{0.2101}                                                    \\ \hline
\multirow{4}{*}{Boston}  & Peak RMSE               & 0.0790                                                                     & 0.0653                                                                      & 0.0385                                                             & \textbf{0.0362}                                                    \\
                         & Peak MAE                & 0.0550                                                                     & 0.0481                                                                      & 0.0288                                                             & \textbf{0.0273}                                                    \\
                         & Peak Mean Signed Error  & -0.0427                                                                    & -0.0260                                                                     & -0.0086                                                            & \textbf{-0.0060}                                                   \\
                         & Peak Max Absolute Error & 0.4446                                                                     & 0.4149                                                                      & 0.2050                                                             & \textbf{0.1881}                                                    \\ \hline
\multirow{4}{*}{CBBT}    & Peak RMSE               & 0.0922                                                                     & 0.0670                                                                      & 0.0360                                                             & \textbf{0.0357}                                                    \\
                         & Peak MAE                & 0.0753                                                                     & 0.0548                                                                      & \textbf{0.0282}                                                    & \textbf{0.0282}                                                    \\
                         & Peak Mean Signed Error  & -0.0609                                                                    & -0.0415                                                                     & -0.0117                                                            & \textbf{-0.0104}                                                   \\
                         & Peak Max Absolute Error & 0.2959                                                                     & 0.2225                                                                      & 0.1446                                                             & \textbf{0.1332}                                                    \\ \hline
\multirow{4}{*}{Lewes}   & Peak RMSE               & 0.0972                                                                     & 0.0764                                                                      & 0.0391                                                             & \textbf{0.0375}                                                    \\
                         & Peak MAE                & 0.0734                                                                     & 0.0583                                                                      & 0.0287                                                             & \textbf{0.0279}                                                    \\
                         & Peak Mean Signed Error  & -0.0588                                                                    & -0.0416                                                                     & -0.0075                                                            & \textbf{-0.0038}                                                   \\
                         & Peak Max Absolute Error & 0.4281                                                                     & 0.3044                                                                      & \textbf{0.1792}                                                    & 0.1849                                                            
\end{tabular}
}
\end{table}

\begin{table}[!ht]
\centering
\caption{1\% Peak time performance comparison between our proposed PACT and baseline model on NCEP Reanalysis dataset at four different stations. Each station is trained and evaluated independently. Numbers are in meters.\label{table:1-peak}}
\scalebox{0.6}{
\begin{tabular}{cccccc}
Stations                 & Metrics                 & \begin{tabular}[c]{@{}c@{}}ST-GNN Baseline\\ (0h Historical Forcing)\end{tabular} & \begin{tabular}[c]{@{}c@{}}ST-GNN Baseline\\ (12h Historical Forcing)\end{tabular} & \begin{tabular}[c]{@{}c@{}}PACT\\ (Base Config, Ours)\end{tabular} & \begin{tabular}[c]{@{}c@{}}PACT\\ (Best Config, Ours)\end{tabular} \\ \hline
\multirow{4}{*}{Battery} & Peak RMSE               & 0.2151                                                                     & 0.1307                                                                      & 0.0627                                                             & \textbf{0.0556}                                                    \\
                         & Peak MAE                & 0.1925                                                                     & 0.1066                                                                      & 0.0505                                                             & \textbf{0.0438}                                                    \\
                         & Peak Mean Signed Error  & -0.1888                                                                    & -0.0983                                                                     & -0.0260                                                            & \textbf{-0.0193}                                                   \\
                         & Peak Max Absolute Error & 0.4228                                                                     & 0.3206                                                                      & 0.1718                                                             & \textbf{0.1637}                                                    \\ \hline
\multirow{4}{*}{Boston}  & Peak RMSE               & 0.1802                                                                     & 0.1289                                                                      & 0.0588                                                             & \textbf{0.0523}                                                    \\
                         & Peak MAE                & 0.1613                                                                     & 0.1030                                                                      & 0.0461                                                             & \textbf{0.0405}                                                    \\
                         & Peak Mean Signed Error  & -0.1606                                                                    & -0.0985                                                                     & -0.0263                                                            & \textbf{-0.0205}                                                   \\
                         & Peak Max Absolute Error & 0.4446                                                                     & 0.4149                                                                      & 0.2050                                                             & \textbf{0.1881}                                                    \\ \hline
\multirow{4}{*}{CBBT}    & Peak RMSE               & 0.1500                                                                     & 0.0880                                                                      & 0.0441                                                             & \textbf{0.0439}                                                    \\
                         & Peak MAE                & 0.1332                                                                     & 0.0745                                                                      & 0.0338                                                             & \textbf{0.0331}                                                    \\
                         & Peak Mean Signed Error  & -0.1258                                                                    & -0.0635                                                                     & \textbf{-0.0250}                                                   & -0.0274                                                            \\
                         & Peak Max Absolute Error & 0.2959                                                                     & 0.2051                                                                      & 0.1446                                                             & \textbf{0.1265}                                                    \\ \hline
\multirow{4}{*}{Lewes}   & Peak RMSE               & 0.1916                                                                     & 0.1205                                                                      & 0.0626                                                             & \textbf{0.0603}                                                    \\
                         & Peak MAE                & 0.1674                                                                     & 0.0966                                                                      & 0.0488                                                             & \textbf{0.0468}                                                    \\
                         & Peak Mean Signed Error  & -0.1648                                                                    & -0.0880                                                                     & -0.0244                                                            & \textbf{-0.0227}                                                   \\
                         & Peak Max Absolute Error & 0.4281                                                                     & 0.3044                                                                      & \textbf{0.1792}                                                    & 0.1849                                                            
\end{tabular}
}
\end{table}

To complement the top 5\% peak-evaluation results reported in the main content, this appendix provides additional results using stricter top 1\% and broader top 10\% peak thresholds. The top 1\% setting focuses on the rarest extreme-surge events, while the top 10\% setting evaluates model performance over a wider range of high-surge conditions. All models are trained and evaluated independently for each station using the same protocol as in the main experiments. These additional threshold-based evaluations serve as a sensitivity analysis for the peak-prediction conclusions discussed in the main text. The corresponding results are summarized in Table~\ref{table:1-peak} and Table~\ref{table:10-peak}.

\section{Additional Mean Absolute Error Results for Cross-Dataset Evaluation}
\label{appendix:additional-cross-dataset}

To complement the RMSE results of cross-dataset evaluation in the main content, this appendix provides additional MAE results, summarized in Table~\ref{table:cross-dataset-past-mae} and Table~\ref{table:cross-dataset-future-mae}.

\begin{table}[!t]
\centering
\caption{Past-only MAE error cross-dataset evaluation results of our proposed PACT with the best configuration, trained on one dataset and evaluated on other datasets. Numbers are in meters.\label{table:cross-dataset-past-mae}}
\setlength{\tabcolsep}{6pt}
\renewcommand{\arraystretch}{1.15}
\begin{tabular}{c c c c c c c}
& \multicolumn{5}{c}{Cross-Model Evaluation Dataset} \\
\cmidrule(lr){2-7}
\textbf{Train/Val} & NCEP & AWI & CNRM & EC\_EARTH & MPI & MRI \\
\midrule
NCEP      & / & 0.1329 & 0.1500 & 0.1130 & 0.1315 & 0.1475 \\
AWI       & 0.0410 & / & 0.0437 & 0.0395 & 0.0310 & 0.0375 \\
CNRM      & 0.0383 & 0.0745 & / & 0.0994 & 0.0758 & 0.0456 \\
EC\_EARTH &  & 0.0353 & 0.0615 & / & 0.0374 & 0.0494 \\
MPI       & 0.0613 & 0.0305 & 0.0576 & 0.0431 & / & 0.0423 \\
MRI       & 0.0489 & 0.0355 & 0.0336 & 0.0590 & 0.0378  & / \\
\end{tabular}
\end{table}

\begin{table}[!t]
\centering
\caption{Future-period MAE error cross-dataset evaluation results of our proposed PACT with best configuration, trained on one dataset and evaluated on other datasets. "n.a." is used for evaluations on NCEP as it only includes past years. Numbers are in meters.\label{table:cross-dataset-future-mae}}
\setlength{\tabcolsep}{6pt}
\renewcommand{\arraystretch}{1.15}
\begin{tabular}{c c c c c c c}
& \multicolumn{5}{c}{Cross-Model Evaluation Dataset} \\
\cmidrule(lr){2-7}
\textbf{Train/Val} & NCEP & AWI & CNRM & EC\_EARTH & MPI & MRI \\
\midrule
NCEP      & / & 0.1234 & 0.1444 & 0.1093 & 0.1289 & 0.1467 \\
AWI       & n.a. & / & 0.0404 & 0.0409 & 0.0314 & 0.0369 \\
CNRM      & n.a. & 0.0691 & / & 0.0952 & 0.0724 & 0.0420 \\
EC\_EARTH & n.a. & 0.0355 & 0.0588 & / & 0.0377 & 0.0511 \\
MPI       & n.a. & 0.0301 & 0.0547 & 0.0432 & / & 0.0422 \\
MRI       & n.a. & 0.0346 & 0.0318 & 0.0588 & 0.0386 & / \\
\end{tabular}
\end{table}

\acknowledgments
This work is supported by the National Science Foundation under grant \#2423211. The authors acknowledge the Texas Advanced Computing Center (TACC) at The University of Texas at Austin for providing computational resources that have contributed to the research results reported within this paper. This work also used Jetstream2 at Indiana University through allocation CIS250588 from the Advanced Cyberinfrastructure Coordination Ecosystem: Services \& Support (ACCESS) program, which is supported by National Science Foundation grants \#2138259, \#2138286, \#2138307, \#2137603, and \#2138296


%
\bibliography{main}

@InProceedings{perceiver,
  title = 	 {Perceiver: General Perception with Iterative Attention},
  author =       {Jaegle, Andrew and Gimeno, Felix and Brock, Andy and Vinyals, Oriol and Zisserman, Andrew and Carreira, Joao},
  booktitle = 	 {Proceedings of the 38th International Conference on Machine Learning},
  pages = 	 {4651--4664},
  year = 	 {2021},
  editor = 	 {Meila, Marina and Zhang, Tong},
  volume = 	 {139},
  series = 	 {Proceedings of Machine Learning Research},
  month = 	 {18--24 Jul},
  publisher =    {PMLR},
  url = 	 {https://proceedings.mlr.press/v139/jaegle21a.html}
}

@inproceedings{Hamilton_2017_GraphSAGE,
     author = {Hamilton, William L. and Ying, Rex and Leskovec, Jure},
     title = {Inductive Representation Learning on Large Graphs},
     booktitle = {NIPS},
     year = {2017}
   }

@misc{Liu_2025_GRIT_LP,
      title={GRIT-LP: Graph Transformer with Long-Range Skip Connection and Partitioned Spatial Graphs for Accurate Ice Layer Thickness Prediction}, 
      author={Zesheng Liu and Maryam Rahnemoonfar},
      year={2025},
      eprint={2511.18716},
      archivePrefix={arXiv},
      primaryClass={cs.LG},
      url={https://arxiv.org/abs/2511.18716}, 
}

@inproceedings{jaegle2022perceiverIO,
title={Perceiver {IO}: A General Architecture for Structured Inputs \& Outputs},
author={Andrew Jaegle and Sebastian Borgeaud and Jean-Baptiste Alayrac and Carl Doersch and Catalin Ionescu and David Ding and Skanda Koppula and Daniel Zoran and Andrew Brock and Evan Shelhamer and Olivier J Henaff and Matthew Botvinick and Andrew Zisserman and Oriol Vinyals and Joao Carreira},
booktitle={International Conference on Learning Representations},
year={2022},
url={https://openreview.net/forum?id=fILj7WpI-g}
}

@inproceedings{Vaswani_2017_Attention,
author = {Vaswani, Ashish and Shazeer, Noam and Parmar, Niki and Uszkoreit, Jakob and Jones, Llion and Gomez, Aidan N. and Kaiser, \L{}ukasz and Polosukhin, Illia},
title = {Attention is all you need},
year = {2017},
isbn = {9781510860964},
publisher = {Curran Associates Inc.},
address = {Red Hook, NY, USA},
booktitle = {Proceedings of the 31st International Conference on Neural Information Processing Systems},
pages = {6000–6010},
numpages = {11},
location = {Long Beach, California, USA},
series = {NIPS'17}
}

@inproceedings{NEURIPS2019_MultiHead,
 author = {Michel, Paul and Levy, Omer and Neubig, Graham},
 booktitle = {Advances in Neural Information Processing Systems},
 editor = {H. Wallach and H. Larochelle and A. Beygelzimer and F. d\textquotesingle Alch\'{e}-Buc and E. Fox and R. Garnett},
 pages = {},
 publisher = {Curran Associates, Inc.},
 title = {Are Sixteen Heads Really Better than One?},
 url = {https://proceedings.neurips.cc/paper_files/paper/2019/file/2c601ad9d2ff9bc8b282670cdd54f69f-Paper.pdf},
 volume = {32},
 year = {2019}
}

@inproceedings{voita-etal-2019-analyzing,
    title = "Analyzing Multi-Head Self-Attention: Specialized Heads Do the Heavy Lifting, the Rest Can Be Pruned",
    author = "Voita, Elena  and
      Talbot, David  and
      Moiseev, Fedor  and
      Sennrich, Rico  and
      Titov, Ivan",
    booktitle = "Proceedings of the 57th Annual Meeting of the Association for Computational Linguistics",
    month = jul,
    year = "2019",
    address = "Florence, Italy",
    publisher = "Association for Computational Linguistics",
    url = "https://www.aclweb.org/anthology/P19-1580",
    pages = "5797--5808",
}

@inproceedings{shazeer2017,
title={Outrageously Large Neural Networks: The Sparsely-Gated Mixture-of-Experts Layer},
author={Noam Shazeer and *Azalia Mirhoseini and *Krzysztof Maziarz and Andy Davis and Quoc Le and Geoffrey Hinton and Jeff Dean},
booktitle={International Conference on Learning Representations},
year={2017},
url={https://openreview.net/forum?id=B1ckMDqlg}
}

@inproceedings{Royer2023RevisitingSM,
  title={Revisiting single-gated Mixtures of Experts},
  author={Am{\'e}lie Royer and Ilia Karmanov and Andrii Skliar and Babak Ehteshami Bejnordi and Tijmen Blankevoort},
  booktitle={British Machine Vision Conference},
  year={2023},
  url={https://api.semanticscholar.org/CorpusID:256903747}
}

@InProceedings{Barron_2019_CVPR,
author = {Barron, Jonathan T.},
title = {A General and Adaptive Robust Loss Function},
booktitle = {Proceedings of the IEEE/CVF Conference on Computer Vision and Pattern Recognition (CVPR)},
month = {June},
year = {2019}
}

@article{sebastian2014,
title = {Characterizing hurricane storm surge behavior in Galveston Bay using the SWAN+ADCIRC model},
journal = {Coastal Engineering},
volume = {88},
pages = {171-181},
year = {2014},
issn = {0378-3839},
doi = {https://doi.org/10.1016/j.coastaleng.2014.03.002},
url = {https://www.sciencedirect.com/science/article/pii/S0378383914000556},
author = {Antonia Sebastian and Jennifer Proft and J. Casey Dietrich and Wei Du and Philip B. Bedient and Clint N. Dawson}
}

@article{glahn2009,
  title={The role of the SLOSH model in National Weather Service storm surge forecasting},
  author={Glahn, Bob and Taylor, Arthur and Kurkowski, Nicole and Shaffer, Wilson A},
  journal={National Weather Digest},
  volume={33},
  number={1},
  pages={3--14},
  year={2009}
}

@article{begmohammadi2025,
author = {Begmohammadi, Amirhosein and Lin, Ning and Xi, Dazhi and Blackshaw, Christine},
title = {Assessing Future Coastal Flood Hazards From Tropical Cyclones in the Northeastern United States},
journal = {Earth's Future},
volume = {13},
number = {11},
pages = {e2025EF006063},
doi = {https://doi.org/10.1029/2025EF006063},
url = {https://agupubs.onlinelibrary.wiley.com/doi/abs/10.1029/2025EF006063},
eprint = {https://agupubs.onlinelibrary.wiley.com/doi/pdf/10.1029/2025EF006063},
note = {e2025EF006063 2025EF006063},
year = {2025}
}

@article {butler2012,
      author = "T. Butler and M. U. Altaf and C. Dawson and I. Hoteit and X. Luo and T. Mayo",
      title = "Data Assimilation within the Advanced Circulation (ADCIRC) Modeling Framework for Hurricane Storm Surge Forecasting",
      journal = "Monthly Weather Review",
      year = "2012",
      publisher = "American Meteorological Society",
      address = "Boston MA, USA",
      volume = "140",
      number = "7",
      doi = "10.1175/MWR-D-11-00118.1",
      pages=      "2215 - 2231",
      url = "https://journals.ametsoc.org/view/journals/mwre/140/7/mwr-d-11-00118.1.xml"
}

@article{pachev2023,
title = {A framework for flexible peak storm surge prediction},
journal = {Coastal Engineering},
volume = {186},
pages = {104406},
year = {2023},
issn = {0378-3839},
doi = {https://doi.org/10.1016/j.coastaleng.2023.104406},
url = {https://www.sciencedirect.com/science/article/pii/S0378383923001308},
author = {Benjamin Pachev and Prateek Arora and Carlos del-Castillo-Negrete and Eirik Valseth and Clint Dawson}
}

@article{adcirc,
author = {Luettich, Jr, Richard and Westerink, J. and Scheffner, Norman},
year = {1992},
month = {11},
pages = {143},
title = {ADCIRC: An Advanced Three-Dimensional Circulation Model for Shelves, Coasts, and Estuaries. Report 1. Theory and Methodology of ADCIRC-2DDI and ADCIRC-3DL},
journal = {Dredging Research Program Tech. Rep. DRP-92-6}
}

@article{lin2019,
title = {Storm surge return levels induced by mid-to-late-twenty-first-century extratropical cyclones in the Northeastern United States},
journal = {Climatic Change},
volume = {154},
year = {2019},
doi = {https://doi.org/10.1007/s10584-019-02431-8},
author = {Ning Lin and Reza Marsooli and Brian A. Colle}
}

@article{valle2021,
author = {Ramos-Valle, Alexandra N. and Curchitser, Enrique N. and Bruyère, Cindy L. and McOwen, Sean},
title = {Implementation of an Artificial Neural Network for Storm Surge Forecasting},
journal = {Journal of Geophysical Research: Atmospheres},
volume = {126},
number = {13},
pages = {e2020JD033266},
doi = {https://doi.org/10.1029/2020JD033266},
url = {https://agupubs.onlinelibrary.wiley.com/doi/abs/10.1029/2020JD033266},
eprint = {https://agupubs.onlinelibrary.wiley.com/doi/pdf/10.1029/2020JD033266},
note = {e2020JD033266 2020JD033266},
year = {2021}
}

@article{lockwood2022,
author = {Lockwood, Joseph W. and Lin, Ning and Oppenheimer, Michael and Lai, Ching-Yao},
title = {Using Neural Networks to Predict Hurricane Storm Surge and to Assess the Sensitivity of Surge to Storm Characteristics},
journal = {Journal of Geophysical Research: Atmospheres},
volume = {127},
number = {24},
pages = {e2022JD037617},
doi = {https://doi.org/10.1029/2022JD037617},
url = {https://agupubs.onlinelibrary.wiley.com/doi/abs/10.1029/2022JD037617},
eprint = {https://agupubs.onlinelibrary.wiley.com/doi/pdf/10.1029/2022JD037617},
note = {e2022JD037617 2022JD037617},
year = {2022}
}

@article{tiggeloven2021,
author = {Timothy Tiggeloven and Anaïs Couasnon and Chiem van Straaten and Sanne Muis and Philip Ward},
title = {Exploring deep learning capabilities for surge predictions in coastal areas}, 
journal={Scientific Reports},
year = {2021}}

@article{tadesse2020,
author = {Tadesse, Michael Getachew and Wahl, Thomas and Cid, Alba},
year = {2020},
month = {04},
pages = {},
title = {Data-Driven Modeling of Global Storm Surges},
volume = {7},
journal = {Frontiers in Marine Science},
doi = {10.3389/fmars.2020.00260}
}

@article{sahoo2019,
author = {Sahoo, Bishnupriya},
year = {2019},
month = {12},
pages = {},
title = {Prediction of storm surge and inundation using climatological datasets for the Indian coast using soft computing techniques},
volume = {23},
journal = {Soft Computing},
doi = {10.1007/s00500-019-03775-0}
}

@article{hashemi2016,
author = {M. Reza Hashemi and Malcolm L. Spaulding and Alex Shaw and Hamed Farhadi and Matt Lewis},
year = {2016},
month = {02},
title = {An efficient artificial intelligence model for prediction of tropical storm surge},
volume = {82},
journal = {Natural Hazards}
}

@article{fleming2008,
author = {Fleming, Jason and Fulcher, Crystal and Luettich, Jr, Richard and Estrade, Brett and Allen, Gabrielle and Winer, Harley},
year = {2008},
month = {08},
pages = {},
title = {A real time storm surge forecasting system using ADCIRC},
journal = {Estuarine and Coastal Modeling X},
doi = {10.1061/40990(324)48}
}

@ARTICLE{ncep,
       author = {{Kalnay}, E. and {Kanamitsu}, M. and {Kistler}, R. and {Collins}, W. and {Deaven}, D. and {Gandin}, L. and {Iredell}, M. and {Saha}, S. and {White}, G. and {Woollen}, J. and {Zhu}, Y. and {Leetmaa}, A. and {Reynolds}, B. and {Chelliah}, M. and {Ebisuzaki}, W. and {Higgins}, W. and {Janowiak}, J. and {Mo}, K.~C. and {Ropelewski}, C. and {Wang}, J. and {Jenne}, Roy and {Joseph}, Dennis},
        title = "{The NCEP/NCAR 40-Year Reanalysis Project.}",
      journal = {Bulletin of the American Meteorological Society},
         year = 1996,
        month = mar,
       volume = {77},
       number = {3},
        pages = {437-472},
      note = {Provided by the NOAA PSL, Boulder, Colorado, USA, from their website at https://psl.noaa.gov}
}

@Article{Eyring2016,
AUTHOR = {Eyring, V. and Bony, S. and Meehl, G. A. and Senior, C. A. and Stevens, B. and Stouffer, R. J. and Taylor, K. E.},
TITLE = {Overview of the Coupled Model Intercomparison Project Phase 6 (CMIP6)
experimental design and organization},
JOURNAL = {Geoscientific Model Development},
VOLUME = {9},
YEAR = {2016},
NUMBER = {5},
PAGES = {1937--1958},
URL = {https://gmd.copernicus.org/articles/9/1937/2016/},
DOI = {10.5194/gmd-9-1937-2016}
}

@article {tpxo9,
      author = "Gary D.  Egbert and Svetlana Y.  Erofeeva",
      title = "Efficient Inverse Modeling of Barotropic Ocean Tides",
      journal = "Journal of Atmospheric and Oceanic Technology",
      year = "2002",
      publisher = "American Meteorological Society",
      address = "Boston MA, USA",
      volume = "19",
      number = "2",
      doi = "10.1175/1520-0426(2002)019<0183:EIMOBO>2.0.CO;2",
      pages=      "183 - 204",
      url = "https://journals.ametsoc.org/view/journals/atot/19/2/1520-0426_2002_019_0183_eimobo_2_0_co_2.xml"
}

@article{return_level1,
author = {Lin, N. and Emanuel, K. A. and Smith, J. A. and Vanmarcke, E.},
title = {Risk assessment of hurricane storm surge for New York City},
journal = {Journal of Geophysical Research: Atmospheres},
volume = {115},
number = {D18},
pages = {},
doi = {https://doi.org/10.1029/2009JD013630},
url = {https://agupubs.onlinelibrary.wiley.com/doi/abs/10.1029/2009JD013630},
eprint = {https://agupubs.onlinelibrary.wiley.com/doi/pdf/10.1029/2009JD013630},
year = {2010}
}

@article {mohammad2024,
      author = "Mohammad Ahmadi Gharehtoragh and David Johnson",
      title = "Using surrogate modeling to predict storm surge on evolving landscapes under climate change",
      journal = "npj Natural Hazards",
      year = "2024",
      volume = "1",
      number = "33",
      url = "https://doi.org/10.1038/s44304-024-00032-9"
}

@article{zhu2026,
title = {Developing a novel hybrid deep learning model for extratropical storm surge forecasting: A case in the Bohai Sea},
journal = {Continental Shelf Research},
volume = {297},
pages = {105621},
year = {2026},
issn = {0278-4343},
doi = {https://doi.org/10.1016/j.csr.2025.105621},
url = {https://www.sciencedirect.com/science/article/pii/S0278434325002213},
author = {Zhicheng Zhu and Chengqing Ruan and Qinrong Liu and Zhifeng Wang and Jinsheng Qi}
}

@article{dinapoli2025,
author = {Dinapoli, Matías G. and Simionato, Claudia G.},
title = {On the Impact of Southeastern Pacific-Generated Storm Surges on the Southwestern Atlantic Continental Shelf: Interoceanic Connections Through Coastally Trapped Waves},
journal = {Journal of Geophysical Research: Oceans},
volume = {130},
number = {4},
pages = {e2024JC021685},
doi = {https://doi.org/10.1029/2024JC021685},
url = {https://agupubs.onlinelibrary.wiley.com/doi/abs/10.1029/2024JC021685},
eprint = {https://agupubs.onlinelibrary.wiley.com/doi/pdf/10.1029/2024JC021685},
note = {e2024JC021685 2024JC021685},
year = {2025}
}

@article {catalano2018,
      author = "Arielle J. Catalano and Anthony J. Broccoli",
      title = "Synoptic Characteristics of Surge-Producing Extratropical Cyclones along the Northeast Coast of the United States",
      journal = "Journal of Applied Meteorology and Climatology",
      year = "2018",
      publisher = "American Meteorological Society",
      address = "Boston MA, USA",
      volume = "57",
      number = "1",
      doi = "10.1175/JAMC-D-17-0123.1",
      pages=      "171 - 184",
      url = "https://journals.ametsoc.org/view/journals/apme/57/1/jamc-d-17-0123.1.xml"
}

@article{marsooli2018,
author = {Marsooli, Reza and Lin, Ning},
title = {Numerical Modeling of Historical Storm Tides and Waves and Their Interactions Along the U.S. East and Gulf Coasts},
journal = {Journal of Geophysical Research: Oceans},
volume = {123},
number = {5},
pages = {3844-3874},
doi = {https://doi.org/10.1029/2017JC013434},
url = {https://agupubs.onlinelibrary.wiley.com/doi/abs/10.1029/2017JC013434},
eprint = {https://agupubs.onlinelibrary.wiley.com/doi/pdf/10.1029/2017JC013434},
year = {2018}
}

@article{ayyad2022,
  title={Machine learning-based assessment of storm surge in the New York metropolitan area},
  author={Ayyad, Mahmoud and Hajj, Muhammad R and Marsooli, Reza},
  journal={Scientific Reports},
  volume={12},
  number={1},
  pages={19215},
  year={2022},
  publisher={Nature Publishing Group UK London}
}

@article{KYPRIOTI2023104231,
title = {Spatio-temporal storm surge emulation using Gaussian Process techniques},
journal = {Coastal Engineering},
volume = {180},
pages = {104231},
year = {2023},
issn = {0378-3839},
doi = {https://doi.org/10.1016/j.coastaleng.2022.104231},
url = {https://www.sciencedirect.com/science/article/pii/S0378383922001442},
author = {Aikaterini P. Kyprioti and Christopher Irwin and Alexandros A. Taflanidis and Norberto C. Nadal-Caraballo and Madison C. Yawn and Luke A. Aucoin}
}

@Article{Adeli2023,
author={Adeli, Ehsan
and Sun, Luning
and Wang, Jianxun
and Taflanidis, Alexandros A.},
title={An advanced spatio-temporal convolutional recurrent neural network for storm surge predictions},
journal={Neural Computing and Applications},
year={2023},
month={Sep},
day={01},
volume={35},
number={26},
pages={18971-18987},
issn={1433-3058},
doi={10.1007/s00521-023-08719-2},
url={https://doi.org/10.1007/s00521-023-08719-2}
}

@article{LEE2021104024,
title = {Rapid prediction of peak storm surge from tropical cyclone track time series using machine learning},
journal = {Coastal Engineering},
volume = {170},
pages = {104024},
year = {2021},
issn = {0378-3839},
doi = {https://doi.org/10.1016/j.coastaleng.2021.104024},
url = {https://www.sciencedirect.com/science/article/pii/S0378383921001691},
author = {Jun-Whan Lee and Jennifer L. Irish and Michelle T. Bensi and Douglas C. Marcy}
}

@article{XIE2023102179,
title = {Developing a deep learning-based storm surge forecasting model},
journal = {Ocean Modelling},
volume = {182},
pages = {102179},
year = {2023},
issn = {1463-5003},
doi = {https://doi.org/10.1016/j.ocemod.2023.102179},
url = {https://www.sciencedirect.com/science/article/pii/S1463500323000203},
author = {Wenhong Xie and Guangjun Xu and Hongchun Zhang and Changming Dong}
}

@Article{Wang2023,
author={Wang, Tiantian
and Liu, Tiezhong
and Lu, Yunmeng},
title={A hybrid multi-step storm surge forecasting model using multiple feature selection, deep learning neural network and transfer learning},
journal={Soft Computing},
year={2023},
month={Jan},
day={01},
volume={27},
number={2},
pages={935-952},
issn={1433-7479},
doi={10.1007/s00500-022-07508-8},
url={https://doi.org/10.1007/s00500-022-07508-8}
}

@article{KIM2019101871,
title = {Artificial neural network-based storm surge forecast model: Practical application to Sakai Minato, Japan},
journal = {Applied Ocean Research},
volume = {91},
pages = {101871},
year = {2019},
issn = {0141-1187},
doi = {https://doi.org/10.1016/j.apor.2019.101871},
url = {https://www.sciencedirect.com/science/article/pii/S014111871830899X},
author = {Sooyoul Kim and Shunqi Pan and Hajime Mase}
}

@Article{w12092394,
AUTHOR = {Chao, Wei-Ting and Young, Chih-Chieh and Hsu, Tai-Wen and Liu, Wen-Cheng and Liu, Chian-Yi},
TITLE = {Long-Lead-Time Prediction of Storm Surge Using Artificial Neural Networks and Effective Typhoon Parameters: Revisit and Deeper Insight},
JOURNAL = {Water},
VOLUME = {12},
YEAR = {2020},
NUMBER = {9},
ARTICLE-NUMBER = {2394},
URL = {https://www.mdpi.com/2073-4441/12/9/2394},
ISSN = {2073-4441},
DOI = {10.3390/w12092394}
}

@Article{jmse10121980,
AUTHOR = {Wei, Zhangping and Nguyen, Hai Cong},
TITLE = {Storm Surge Forecast Using an Encoder–Decoder Recurrent Neural Network Model},
JOURNAL = {Journal of Marine Science and Engineering},
VOLUME = {10},
YEAR = {2022},
NUMBER = {12},
ARTICLE-NUMBER = {1980},
URL = {https://www.mdpi.com/2077-1312/10/12/1980},
ISSN = {2077-1312},
DOI = {10.3390/jmse10121980}
}

@article{TEDESCO2024102334,
title = {Bias correction of operational storm surge forecasts using Neural Networks},
journal = {Ocean Modelling},
volume = {188},
pages = {102334},
year = {2024},
issn = {1463-5003},
doi = {https://doi.org/10.1016/j.ocemod.2024.102334},
url = {https://www.sciencedirect.com/science/article/pii/S1463500324000210},
author = {Paulina Tedesco and Jean Rabault and Martin Lilleeng Sætra and Nils Melsom Kristensen and Ole Johan Aarnes and Øyvind Breivik and Cecilie Mauritzen and Øyvind Sætra}
}

@article{JIANG2024104512,
title = {Advancing storm surge forecasting from scarce observation data: A causal-inference based Spatio-Temporal Graph Neural Network approach},
journal = {Coastal Engineering},
volume = {190},
pages = {104512},
year = {2024},
issn = {0378-3839},
doi = {https://doi.org/10.1016/j.coastaleng.2024.104512},
url = {https://www.sciencedirect.com/science/article/pii/S0378383924000607},
author = {Wenjun Jiang and Jize Zhang and Yuerong Li and Dongqin Zhang and Gang Hu and Huanxiang Gao and Zhongdong Duan}
}

@misc{kristensen2026flodatadrivenlimitedareastorm,
      title={Flo: A data-driven limited-area storm surge model}, 
      author={Nils Melsom Kristensen and Mateusz Matuszak and Paulina Tedesco and Ina Kristine Berentsen Kullmann and Johannes Röhrs},
      year={2026},
      eprint={2601.02090},
      archivePrefix={arXiv},
      primaryClass={physics.ao-ph},
      url={https://arxiv.org/abs/2601.02090}, 
}

@article{awi,
  title={Simulations for CMIP6 with the AWI climate model AWI-CM-1-1},
  author={Semmler, Tido and Danilov, Sergey and Gierz, Paul and Goessling, Helge F and Hegewald, Jan and Hinrichs, Claudia and Koldunov, Nikolay and Khosravi, Narges and Mu, Longjiang and Rackow, Thomas and others},
  journal={Journal of Advances in Modeling Earth Systems},
  volume={12},
  number={9},
  pages={e2019MS002009},
  year={2020},
  publisher={Wiley Online Library}
}

@article{cnrm,
  title={Simulation of observed climate changes in 1850--2014 with climate model INM-CM5},
  author={Volodin, Evgeny and Gritsun, Andrey},
  journal={Earth System Dynamics},
  volume={9},
  number={4},
  pages={1235--1242},
  year={2018},
  publisher={Copernicus GmbH}
}

@article{earth,
  title={Replicability of the EC-Earth3 Earth system model under a change in computing environment},
  author={Massonnet, Fran{\c{c}}ois and M{\'e}n{\'e}goz, Martin and Acosta, Mario and Yepes-Arb{\'o}s, Xavier and Exarchou, Eleftheria and Doblas-Reyes, Francisco J},
  journal={Geoscientific Model Development},
  volume={13},
  number={3},
  pages={1165--1178},
  year={2020},
  publisher={Copernicus GmbH}
}

@article{mpi,
  title={Max planck institute earth system model (MPI-ESM1. 2) for the high-resolution model intercomparison project (HighResMIP)},
  author={Gutjahr, Oliver and Putrasahan, Dian and Lohmann, Katja and Jungclaus, Johann H and von Storch, Jin-Song and Br{\"u}ggemann, Nils and Haak, Helmuth and St{\"o}ssel, Achim},
  journal={Geoscientific Model Development},
  volume={12},
  number={7},
  pages={3241--3281},
  year={2019},
  publisher={Copernicus GmbH}
}

@article{mri,
  title={The Meteorological Research Institute Earth System Model version 2.0, MRI-ESM2. 0: Description and basic evaluation of the physical component},
  author={Yukimoto, Seiji and Kawai, Hideaki and Koshiro, Tsuyoshi and Oshima, Naga and Yoshida, Kohei and Urakawa, Shogo and Tsujino, Hiroyuki and Deushi, Makoto and Tanaka, Taichu and Hosaka, Masahiro and others},
  journal={Journal of the Meteorological Society of Japan. Ser. II},
  volume={97},
  number={5},
  pages={931--965},
  year={2019},
  publisher={Meteorological Society of Japan}
}

@article{Movagha_2025,
author = {Toofani Movaghar, P. and Taflandis, A. A. and Nadal-Caraballo, N. C. and Zhang, J.},
title = {Computationally and Memory Efficient Graph Neural Network for Emulating Storm Surge Inundation State Over Extended Geospatial Domains},
journal = {Journal of Geophysical Research: Machine Learning and Computation},
volume = {3},
number = {2},
pages = {e2025JH001032},
doi = {https://doi.org/10.1029/2025JH001032},
url = {https://agupubs.onlinelibrary.wiley.com/doi/abs/10.1029/2025JH001032},
eprint = {https://agupubs.onlinelibrary.wiley.com/doi/pdf/10.1029/2025JH001032},
note = {e2025JH001032 2025JH001032},
year = {2026}
}

@article{Storm_Surge_Hazards,
author = {Needham, Hal F. and Keim, Barry D. and Sathiaraj, David},
title = {A review of tropical cyclone-generated storm surges: Global data sources, observations, and impacts},
journal = {Reviews of Geophysics},
volume = {53},
number = {2},
pages = {545-591},
doi = {https://doi.org/10.1002/2014RG000477},
url = {https://agupubs.onlinelibrary.wiley.com/doi/abs/10.1002/2014RG000477},
eprint = {https://agupubs.onlinelibrary.wiley.com/doi/pdf/10.1002/2014RG000477},
year = {2015}
}

@article{Bertin_StormSurge,
author = {Xavier Bertin},
title = {Storm surges and coastal flooding: status and challenges},
journal = {La Houille Blanche},
volume = {102},
number = {2},
pages = {64--70},
year = {2016},
publisher = {Taylor \& Francis},
doi = {10.1051/lhb/2016020},


URL = { 
    
        https://doi.org/10.1051/lhb/2016020
    
    

},
eprint = { 
    
        https://doi.org/10.1051/lhb/2016020
    
    

}

}

@article{Qian_2024,
author = {Qian, Xiaojuan and Hwang, Sooncheol and Son, Sangyoung},
title = {A Study on Key Determinants in Enhancing Storm Surges Along the Coast: Interplay Between Tropical Cyclone Motion and Coastal Geometry},
journal = {Journal of Geophysical Research: Oceans},
volume = {129},
number = {2},
pages = {e2023JC020400},
doi = {https://doi.org/10.1029/2023JC020400},
url = {https://agupubs.onlinelibrary.wiley.com/doi/abs/10.1029/2023JC020400},
eprint = {https://agupubs.onlinelibrary.wiley.com/doi/pdf/10.1029/2023JC020400},
note = {e2023JC020400 2023JC020400},
year = {2024}
}

@Article{Islam2021,
author={Islam, Md. Rezuanul
and Lee, Chia-Ying
and Mandli, Kyle T.
and Takagi, Hiroshi},
title={A new tropical cyclone surge index incorporating the effects of coastal geometry, bathymetry and storm information},
journal={Scientific Reports},
year={2021},
month={Aug},
day={18},
volume={11},
number={1},
pages={16747},
issn={2045-2322},
doi={10.1038/s41598-021-95825-7},
url={https://doi.org/10.1038/s41598-021-95825-7}
}

@article{DIETRICH_2011_SWAN_ADCIRC,
title = {Modeling hurricane waves and storm surge using integrally-coupled, scalable computations},
journal = {Coastal Engineering},
volume = {58},
number = {1},
pages = {45-65},
year = {2011},
issn = {0378-3839},
doi = {https://doi.org/10.1016/j.coastaleng.2010.08.001},
url = {https://www.sciencedirect.com/science/article/pii/S0378383910001250},
author = {J.C. Dietrich and M. Zijlema and J.J. Westerink and L.H. Holthuijsen and C. Dawson and R.A. Luettich and R.E. Jensen and J.M. Smith and G.S. Stelling and G.W. Stone}
}

@Article{Dietrich_2012_SWAN_ADCIRC,
author={Dietrich, J. C.
and Tanaka, S.
and Westerink, J. J.
and Dawson, C. N.
and Luettich, R. A.
and Zijlema, M.
and Holthuijsen, L. H.
and Smith, J. M.
and Westerink, L. G.
and Westerink, H. J.},
title={Performance of the Unstructured-Mesh, SWAN+ADCIRC Model in Computing Hurricane Waves and Surge},
journal={Journal of Scientific Computing},
year={2012},
month={Aug},
day={01},
volume={52},
number={2},
pages={468-497},
issn={1573-7691},
doi={10.1007/s10915-011-9555-6},
url={https://doi.org/10.1007/s10915-011-9555-6}
}

@article{Pachev_2023,
title = {A framework for flexible peak storm surge prediction},
journal = {Coastal Engineering},
volume = {186},
pages = {104406},
year = {2023},
issn = {0378-3839},
doi = {https://doi.org/10.1016/j.coastaleng.2023.104406},
url = {https://www.sciencedirect.com/science/article/pii/S0378383923001308},
author = {Benjamin Pachev and Prateek Arora and Carlos del-Castillo-Negrete and Eirik Valseth and Clint Dawson}
}

@Article{Dullaart_2023,
AUTHOR = {Dullaart, J. C. M. and Muis, S. and de Moel, H. and Ward, P. J. and Eilander, D. and Aerts, J. C. J. H.},
TITLE = {Enabling dynamic modelling of coastal flooding by defining storm tide hydrographs},
JOURNAL = {Natural Hazards and Earth System Sciences},
VOLUME = {23},
YEAR = {2023},
NUMBER = {5},
PAGES = {1847--1862},
URL = {https://nhess.copernicus.org/articles/23/1847/2023/},
DOI = {10.5194/nhess-23-1847-2023}
}

@article{Bernier_2006,
author = {Bernier, N. B. and Thompson, K. R.},
title = {Predicting the frequency of storm surges and extreme sea levels in the northwest Atlantic},
journal = {Journal of Geophysical Research: Oceans},
volume = {111},
number = {C10},
pages = {},
doi = {https://doi.org/10.1029/2005JC003168},
url = {https://agupubs.onlinelibrary.wiley.com/doi/abs/10.1029/2005JC003168},
eprint = {https://agupubs.onlinelibrary.wiley.com/doi/pdf/10.1029/2005JC003168},
year = {2006}
}

@Article{Muis2016,
author={Muis, Sanne
and Verlaan, Martin
and Winsemius, Hessel C.
and Aerts, Jeroen C. J. H.
and Ward, Philip J.},
title={A global reanalysis of storm surges and extreme sea levels},
journal={Nature Communications},
year={2016},
month={Jun},
day={27},
volume={7},
number={1},
pages={11969},
issn={2041-1723},
doi={10.1038/ncomms11969},
url={https://doi.org/10.1038/ncomms11969}
}

@Article{ONeill_2016_ScenarioMIP,
AUTHOR = {O'Neill, B. C. and Tebaldi, C. and van Vuuren, D. P. and Eyring, V. and Friedlingstein, P. and Hurtt, G. and Knutti, R. and Kriegler, E. and Lamarque, J.-F. and Lowe, J. and Meehl, G. A. and Moss, R. and Riahi, K. and Sanderson, B. M.},
TITLE = {The Scenario Model Intercomparison Project (ScenarioMIP) for CMIP6},
JOURNAL = {Geoscientific Model Development},
VOLUME = {9},
YEAR = {2016},
NUMBER = {9},
PAGES = {3461--3482},
URL = {https://gmd.copernicus.org/articles/9/3461/2016/},
DOI = {10.5194/gmd-9-3461-2016}
}

@article{Muis_2023_CMIP6_StormSurges,
author = {Muis, Sanne and Aerts, Jeroen C. J. H. and Á. Antolínez, José A. and Dullaart, Job C. and Duong, Trang Minh and Erikson, Li and Haarsma, Rein J. and Apecechea, Maialen Irazoqui and Mengel, Matthias and Le Bars, Dewi and O’Neill, Andrea and Ranasinghe, Roshanka and Roberts, Malcolm J. and Verlaan, Martin and Ward, Philip J. and Yan, Kun},
title = {Global Projections of Storm Surges Using High-Resolution CMIP6 Climate Models},
journal = {Earth's Future},
volume = {11},
number = {9},
pages = {e2023EF003479},
doi = {https://doi.org/10.1029/2023EF003479},
url = {https://agupubs.onlinelibrary.wiley.com/doi/abs/10.1029/2023EF003479},
eprint = {https://agupubs.onlinelibrary.wiley.com/doi/pdf/10.1029/2023EF003479},
note = {e2023EF003479 2023EF003479},
year = {2023}
}

@article{Sandeep_ETC_2024,
author = {Chinta, Sandeep and Schlosser, C. Adam and Gao, Xiang and Hodges, Kevin},
title = {Future Changes in Winter-Time Extratropical Cyclones Over South Africa From CORDEX-CORE Simulations},
journal = {Earth's Future},
volume = {13},
number = {1},
pages = {e2024EF005289},
doi = {https://doi.org/10.1029/2024EF005289},
url = {https://agupubs.onlinelibrary.wiley.com/doi/abs/10.1029/2024EF005289},
eprint = {https://agupubs.onlinelibrary.wiley.com/doi/pdf/10.1029/2024EF005289},
note = {e2024EF005289 2024EF005289},
year = {2025}
}

@misc{Nader_2026_StormNet,
      title={StormNet: Improving storm surge predictions with a GNN-based spatio-temporal offset forecasting model}, 
      author={Noujoud Nader and Stefanos Giaremis and Clint Dawson and Carola Kaiser and Karame Mohammadiporshokooh and Hartmut Kaiser},
      year={2026},
      eprint={2604.20688},
      archivePrefix={arXiv},
      primaryClass={cs.LG},
      url={https://arxiv.org/abs/2604.20688}, 
}

@article{rudeva2011,
  title={Composite analysis of North Atlantic extratropical cyclones in NCEP--NCAR reanalysis data},
  author={Rudeva, Irina and Gulev, Sergey K},
  journal={Monthly Weather Review},
  volume={139},
  number={5},
  pages={1419--1446},
  year={2011}
}

@article{colle2010new,
  title={New York City storm surges: Climatology and an analysis of the wind and cyclone evolution},
  author={Colle, Brian A and Rojowsky, Katherine and Buonaito, Frank},
  journal={Journal of Applied Meteorology and Climatology},
  volume={49},
  number={1},
  pages={85--100},
  year={2010}
}

@article{priestley2020,
  title={An overview of the extratropical storm tracks in CMIP6 historical simulations},
  author={Priestley, Matthew DK and Ackerley, Duncan and Catto, Jennifer L and Hodges, Kevin I and McDonald, Ruth E and Lee, Robert W},
  journal={Journal of Climate},
  volume={33},
  number={15},
  pages={6315--6343},
  year={2020}
}

@article{gore2023,
  title={Connecting Large-Scale Meteorological Patterns to Extratropical Cyclones in CMIP6 Climate Models Using Self-Organizing Maps},
  author={Gore, Michelle J and Zarzycki, Colin M and Gervais, Melissa M},
  journal={Earth's Future},
  volume={11},
  number={8},
  pages={e2022EF003211},
  year={2023},
  publisher={Wiley Online Library}
}

@article{longo2026,
  title={A deep learning framework for extreme storm surge modeling under future climate scenarios},
  author={Longo, Emiliano and Ficch{\`\i}, Andrea and Verlaan, Martin and Muis, Sanne and Castelletti, Andrea},
  journal={Earth's Future},
  volume={14},
  number={3},
  pages={e2025EF007072},
  year={2026},
  publisher={Wiley Online Library}
}

@misc{NOAA_COOPS_WaterLevels,
  author       = {{NOAA Center for Operational Oceanographic Products and Services}},
  title        = {{Tides \& Great Lakes Water Levels}},
  year         = {n.d.},
  howpublished = {\url{https://tidesandcurrents.noaa.gov/water_level_info.html}},
}

@article{Koo_Rahnemoonfar_2025, title={Graph convolutional network as a fast statistical emulator for numerical ice sheet modeling}, volume={71}, DOI={10.1017/jog.2024.93}, journal={Journal of Glaciology}, author={Koo, Younghyun and Rahnemoonfar, Maryam}, year={2025}, pages={e15}}

@Article{Koo_Helheim,
AUTHOR = {Koo, Y. and Cheng, G. and Morlighem, M. and Rahnemoonfar, M.},
TITLE = {Calibrating calving parameterizations using graph neural network emulators: application to Helheim Glacier, East Greenland},
JOURNAL = {The Cryosphere},
VOLUME = {19},
YEAR = {2025},
NUMBER = {7},
PAGES = {2583--2599},
URL = {https://tc.copernicus.org/articles/19/2583/2025/},
DOI = {10.5194/tc-19-2583-2025}
}

@INPROCEEDINGS{Zalatan_RadarConf23,
  author={Zalatan, Benjamin and Rahnemoonfar, Maryam},
  booktitle={2023 IEEE Radar Conference (RadarConf23)}, 
  title={Recurrent Graph Convolutional Networks for Spatiotemporal Prediction of Snow Accumulation Using Airborne Radar}, 
  year={2023},
  volume={},
  number={},
  pages={1-6},
  doi={10.1109/RadarConf2351548.2023.10149562}}

@INPROCEEDINGS{Zalatan_IGARSS2023,
  author={Zalatan, Benjamin and Rahnemoonfar, Maryam},
  booktitle={IGARSS 2023 - 2023 IEEE International Geoscience and Remote Sensing Symposium}, 
  title={Prediction of Annual Snow Accumulation Using a Recurrent Graph Convolutional Approach}, 
  year={2023},
  volume={},
  number={},
  pages={5344-5347},
  doi={10.1109/IGARSS52108.2023.10283236}}

@INPROCEEDINGS{Zalatan_ICIP2023,
  author={Zalatan, Benjamin and Rahnemoonfar, Maryam},
  booktitle={2023 IEEE International Conference on Image Processing (ICIP)}, 
  title={Prediction of Deep Ice Layer Thickness Using Adaptive Recurrent Graph Neural Networks}, 
  year={2023},
  volume={},
  number={},
  pages={2835-2839},
  doi={10.1109/ICIP49359.2023.10222391}}

@INPROCEEDINGS{Rahnemoonfar_PIML_IGARSS2024,
  author={Rahnemoonfar, Maryam and Zalatan, Benjamin},
  booktitle={IGARSS 2024 - 2024 IEEE International Geoscience and Remote Sensing Symposium}, 
  title={Physics-informed Machine Learning for Deep Ice Layer Tracing in SAR images}, 
  year={2024},
  volume={},
  number={},
  pages={6938-6942},
  doi={10.1109/IGARSS53475.2024.10641831}}

@misc{liu2024learningspatiotemporalpatternspolar,
      title={Learning Spatio-Temporal Patterns of Polar Ice Layers With Physics-Informed Graph Neural Network}, 
      author={Zesheng Liu and Maryam Rahnemoonfar},
      year={2024},
      eprint={2406.15299},
      archivePrefix={arXiv},
      primaryClass={cs.LG},
      url={https://arxiv.org/abs/2406.15299}, 
}

@misc{liu2024multibranchspatiotemporalgraphneural,
      title={Multi-branch Spatio-Temporal Graph Neural Network For Efficient Ice Layer Thickness Prediction}, 
      author={Zesheng Liu and Maryam Rahnemoonfar},
      year={2024},
      eprint={2411.04055},
      archivePrefix={arXiv},
      primaryClass={cs.LG},
      url={https://arxiv.org/abs/2411.04055}, 
}

@INPROCEEDINGS{Liu_PIML_RADAR25,
  author={Liu, Zesheng and Rahnemoonfar, Maryam},
  booktitle={2025 IEEE International Radar Conference (RADAR)}, 
  title={Physics-Informed Spatio-Temporal Graph Neural Network for Efficient Deep Ice Layer Thickness Estimation in Radar Imagery}, 
  year={2025},
  volume={},
  number={},
  pages={1-6},
  doi={10.1109/RADAR52380.2025.11031955}}

@inproceedings{Liu_SPIE25,
  title={Locate and extend: a geometric deep learning strategy for predicting polar ice layer structures using graph neural networks},
  author={Liu, Zesheng and Rahnemoonfar, Maryam},
  booktitle={Pattern Recognition and Prediction XXXVI},
  volume={13464},
  pages={1346402},
  year={2025},
  organization={SPIE}
}

@INPROCEEDINGS{Liu_GRIT_IGARSS2025,
  author={Liu, Zesheng and Rahnemoonfar, Maryam},
  booktitle={IGARSS 2025 - 2025 IEEE International Geoscience and Remote Sensing Symposium}, 
  title={GRIT: Graph Transformer For Internal Ice Layer Thickness Prediction}, 
  year={2025},
  volume={},
  number={},
  pages={1-5},
  doi={10.1109/IGARSS55030.2025.11243115}}

@INPROCEEDINGS{Liu_STGRIT_ICIP25,
  author={Liu, Zesheng and Rahnemoonfar, Maryam},
  booktitle={2025 IEEE International Conference on Image Processing (ICIP)}, 
  title={ST-GRIT: Spatio-Temporal Graph Transformer For Internal Ice Layer Thickness Prediction}, 
  year={2025},
  volume={},
  number={},
  pages={1109-1114},
  doi={10.1109/ICIP55913.2025.11084445}}

@misc{liu2026kstemitknowledgeinformedspatiotemporalefficient,
      title={K-STEMIT: Knowledge-Informed Spatio-Temporal Efficient Multi-Branch Graph Neural Network for Subsurface Stratigraphy Thickness Estimation from Radar Data}, 
      author={Zesheng Liu and Maryam Rahnemoonfar},
      year={2026},
      eprint={2604.09922},
      archivePrefix={arXiv},
      primaryClass={cs.LG},
      url={https://arxiv.org/abs/2604.09922}, 
}

@misc{liu2025kangcncombiningkolmogorovarnoldnetwork,
      title={KAN-GCN: Combining Kolmogorov-Arnold Network with Graph Convolution Network for an Accurate Ice Sheet Emulator}, 
      author={Zesheng Liu and YoungHyun Koo and Maryam Rahnemoonfar},
      year={2025},
      eprint={2510.24926},
      archivePrefix={arXiv},
      primaryClass={cs.LG},
      url={https://arxiv.org/abs/2510.24926}, 
}
%


%
%
%
%
%

\end{document}